\documentclass{article}
\usepackage[margin=1in]{geometry}
\usepackage{times}

\usepackage{amsmath,amsfonts,bm}

\def\eqref#1{equation~\ref{#1}}

\def\1{\bm{1}}

\DeclareMathAlphabet{\mathsfit}{\encodingdefault}{\sfdefault}{m}{sl}
\SetMathAlphabet{\mathsfit}{bold}{\encodingdefault}{\sfdefault}{bx}{n}

\usepackage{natbib}
\usepackage[utf8]{inputenc} 
\usepackage[T1]{fontenc}    
\usepackage{hyperref} 
\usepackage{amsmath}
\usepackage{amssymb}
\usepackage{mathtools}
\usepackage{amsthm}
\usepackage{subcaption}
\usepackage{caption}

\usepackage{float}
\usepackage{algorithm}
\usepackage{algorithmic}
\usepackage{booktabs}
\usepackage{multirow}
\usepackage{enumitem}
\usepackage{graphicx}
\usepackage{wrapfig}
\usepackage{url}
\usepackage{amsfonts}       
\usepackage{nicefrac}       
\usepackage{microtype}      
\usepackage[table]{xcolor}
\definecolor{motifgreen}{HTML}{88BDA4}
\newcommand{\stddev}[1]{{\scriptsize (#1)}}

\title{iGC: Antibody Optimization via \\ Reinforcement Learning over a Learned \\ Somatic Hypermutation Prior}
\author{
    \textbf{Thao Nguyen$^{1}$\thanks{Work performed during an internship at Genentech.},
    Robert G. Alberstein$^{2}$,
    Ewa Nowara$^{3}$,
    Homa MohammadiPeyhani$^{3}$,} \\
    \textbf{Holly R. Steach$^{2}$,
    Sarah A. Robinson$^{2}$,
    Yuval Shechter$^{2}$ 
    Donald W. Lee$^{2}$,} \\
    \textbf{Jason A. Vander Heiden$^{2}$, 
    Franziska Seeger$^{2}$,
    Andrew Leaver-Fay$^{2}$,
    Richard Bonneau$^{2}$,} \\
    \textbf{Joshua Yao-Yu Lin$^{2}$} \\[2pt]
    \small $^{1}$University of Illinois at Urbana-Champaign \quad
    $^{2}$Genentech \quad
    $^{3}$F. Hoffmann-La Roche Ltd.
}

\begin{document}
\maketitle

\begin{abstract}
Antibody affinity maturation is a natural optimization loop. Individual B cells propose mutations through somatic hypermutation (SHM), variants with higher affinity for antigen are positively selected, and the process repeats cyclically, giving rise to a population of high-affinity antibody producing B cells. A computational model of this process should therefore capture both how mutations arise and how they are selected for antigen-specific affinity. However, modeling these two components jointly is challenging: mutation models capture the sequence patterns of SHM but do not directly optimize antigen binding, while affinity optimization can improve binding without accounting for biologically plausible mutation trajectories.
We introduce \textit{iGC} (\textit{in silico} Germinal Centers), a two-stage reinforcement learning (RL) framework for antigen-specific antibody optimization. Stage~1 learns an SHM mutation prior from parent--child pairs in B-cell lineage trees and domain-adapts it to the target affinity-maturation dataset. Stage~2 initializes an RL policy from this adapted prior and optimizes antigen binding while remaining Kullback–Leibler (KL) regularized toward a frozen copy of the prior. Direct optimization of affinity rewards can lead to degenerate solutions, including excessive use of a small number of highly rewarding substitutions or convergence toward structurally tolerated framework mutations. To mitigate these effects, we introduce a Reward--Tax Meta-Gradient objective. RewardNet smooths the affinity landscape, while TaxNet discourages reward overexploitation; their difference defines a learned meta-reward for policy optimization.
We evaluate whether iGC achieves both objectives: preserving SHM-consistent mutation behavior and improving antigen-specific binding. The mutation prior more than doubles the strongest baseline on mutation-site prediction. Following RL optimization, iGC preserves strong SHM sequence structure, with 57.0\% of predicted mutation positions falling within WRC hotspots (canonical DNA motifs preferentially targeted during SHM). iGC achieves a strong exploration--exploitation balance, yielding competitive reward--mutation-diversity products across candidate-selection cutoffs. These results demonstrate that iGC can jointly model biologically plausible mutation generation and antigen-specific affinity selection, providing an \textit{in silico} framework for iterative antibody affinity maturation. 
\end{abstract}

\section{Introduction}
\begin{figure}[!hbt]
    \centering
    \includegraphics[width=\linewidth]{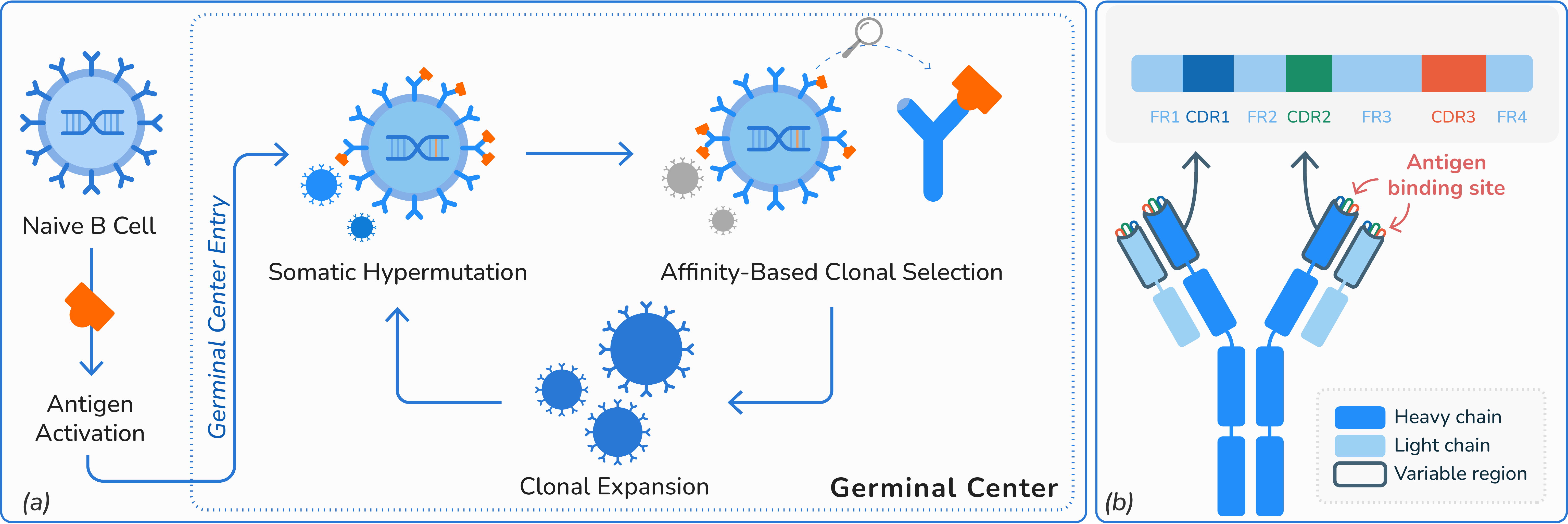}
    \caption{\textit{(a)} Affinity maturation in germinal centers as a repeated mutation--selection process. \textit{(b)} Structure of an antibody and the heavy-chain variable region (VH).}
    \label{fig:gcloop}
\end{figure}

The immune system naturally optimizes antibody sequence repertoires through the process of \emph{affinity maturation} in specialized anatomical structures called germinal centers~\citep{victora2012germinal}. In B cells, somatic hypermutation (SHM) of the antibody-encoding Ig locus introduces mutations concentrated in Complementarity-Determining Regions (CDRs), specifically diversifying antigen binding. Competition for antigen drives selective expansion of high-affinity binders while eliminating sequences that compromise protein structure or exhibit autoimmune cross-reactivity. This can be formally represented as a sequential decision process: a unique antibody sequence represents a state, binding improvement supplies the reward, and each selected mutation alters the epistatic context for subsequent decisions. Iterative cycles of diversification and selection thus function as a natural optimization loop, yielding B cells that express potent, physiologically optimized antibodies (Figure~\ref{fig:gcloop}a).


Computational antibody design typically models only one half of this loop. B-cell-receptor repertoire and antibody language models learn statistical patterns of SHM and natural sequence evolution but do not directly optimize binding to a specified antigen~\citep{ruffolo2021deciphering, olsen2022ablang}. Conversely, antigen-specific affinity-optimization methods search directly for stronger binders but do not explicitly model the sequential, biologically regularized mutation trajectories that produce them~\citep{shan2022deep, luo2022antigen, tagasovska2025property}. Such optimization can move toward sequences that score well for binding yet lie far from mutation paths naturally accessible to the body, bypassing physiological constraints that preserve structural integrity and prevent off-target polyreactivity. Existing approaches also generally optimize individual variants rather than learning how to make successive, context-dependent mutation decisions across the full mutation--selection process.

We introduce \textit{iGC} (\textit{in silico} Germinal Centers), which learns to propose SHM-consistent mutations and select those that improve binding to a specific antigen. iGC separates these roles through two-stage learning. \textbf{Stage~1} first trains a Transformer on parent--child pairs from B-cell lineage trees, producing a mutation prior that captures where SHM acts and which substitutions it favors, and then domain-adapts this prior to the target affinity-maturation dataset. \textbf{Stage~2} initializes a policy from the adapted prior and optimizes antigen-specific binding with REINFORCE~\citep{williams1992simple}. A frozen copy of the adapted SHM prior remains a KL-regularization anchor throughout Stage~2, allowing affinity-driven optimization without unconstrained drift from repertoire-plausible mutation behavior.

To address this problem, we introduce a \textit{Reward--Tax Meta-Gradient} (RTMG) objective. Instead of training the policy directly on raw deep mutational scanning (DMS) rewards, RTMG learns a meta-reward equal to a refined benefit from RewardNet minus a refined exploitation penalty from TaxNet. RewardNet smooths the uneven affinity landscape while preserving agreement with measured binding. TaxNet penalizes recurrent or SHM-accessible sequence outcomes that the optimization can repeatedly exploit. Both networks are learned through their effect on a differentiable probe-policy update, rather than from manually specified targets. The deployed policy is then trained on the detached meta-reward while remaining KL-regularized toward the SHM prior.

The resulting policy improves affinity without discarding SHM structure. After RL, $57.0\%$ of iGC-predicted mutation positions lie in WRC hotspots, compared with $2.0\%$ under random background, and predicted mutability remains correlated with S5F, a statistical model of SHM mutability~\citep{yaari2013models} (Pearson $0.474$). At the same time, iGC generates high-affinity candidates while maintaining mutation diversity, yielding strong reward--mutation-diversity products across candidate-selection cutoffs. To test generalization, we hold out entire germinal-center lineage trees and ask whether iGC can rediscover positive-reward sequences observed only in this unseen test set. Across mutation budgets of two to four mutations, iGC exactly rediscovers 110 of 799 such sequences. Together, these results show that iGC learns diverse, SHM-like mutations that improve antigen-specific binding and recover beneficial sequences from unseen lineages.

In summary, our contributions are:
\begin{itemize}[leftmargin=*]
    \item \textbf{A sequential formulation of antibody optimization.} We model germinal-center affinity maturation as repeated SHM mutation proposals followed by antigen-specific affinity selection.

    \item \textbf{Biologically constrained reinforcement learning.} A learned SHM prior initializes and KL-regularizes antigen-specific policy optimization, coupling realistic mutation proposals with binding-driven selection.

    \item \textbf{Reward--tax meta-gradient optimization.} RewardNet smooths the affinity signal and TaxNet penalizes repeatedly exploitable sequence outcomes, yielding stronger reward--diversity trade-offs while preserving SHM hallmarks.
\end{itemize}


\section{Background}


\paragraph{Antibody Structure and Antigen Recognition.}
Antibodies are Y-shaped proteins made by B lymphocytes that function both as cell-surface receptors and as secreted effector molecules (Figure~\ref{fig:gcloop}b). A single antibody consists of two heavy chains and two light chains, each containing variable and constant domains. Antigen recognition is mediated by the variable domains located at the tips of the antibody, whereas the constant domains are responsible for downstream immune effector functions. Within each heavy- and light-chain variable domain, four structural framework regions (FR1 to FR4) provide structural support, while three complementarity-determining regions (CDR1 to CDR3) form the antigen-binding interface~\citep{chothia1987canonical}. Together, the heavy and light chains contribute six CDRs (CDR-H1 to CDR-H3 and CDR-L1 to CDR-L3) that collectively determine binding specificity and affinity. Among these, CDR-H3 is the most sequence-diverse region and frequently makes the largest contribution to antigen recognition~\citep{chothia1989conformations}.
In this study, we focus exclusively on the heavy-chain variable region (VH), as heavy-chain sequences are more widely available in repertoire datasets and often capture a substantial fraction of antigen-binding determinants, particularly through CDR-H3.

\paragraph{Somatic Hypermutation}
To balance a vast, continuously evolving antigen landscape against physiological constraints, the immune system employs multi-stage antibody adaptation. Initially, stochastic V(D)J recombination generates unique sequences that provide broad, low-affinity coverage against unseen targets, while autoreactive variants are filtered out. Because the immune system cannot pre-optimize for unpredictable targets, these interactions remain sub-optimal. Consequently, repertoire fine-tuning requires \textit{ad hoc} exploration during an active immune response, driven by \textit{somatic hypermutation} (SHM) during germinal center \textit{affinity maturation}~\citep{mishra2018insights}. Given a naive sequence, SHM introduces targeted point mutations into the variable region, and positive selection subsequently expands high-affinity binders. By elevating mutation rates a million-fold, SHM enables a Darwinian evolutionary loop that optimizes antibody binding over days to weeks~\citep{pilzecker2019mutating}, yielding high-affinity populations that mediate lasting immunity.

\section{Related Work}
\textbf{Learning Somatic Hypermutation. } Because SHM follows reproducible sequence-dependent patterns rather than mutating uniformly, many methods aim to learn its mutational grammar from repertoire data~\citep{yaari2013models, tang2022deep, sung2025thrifty, lu2026conditionally}. They differ in the mutational context captured and whether selection is modeled. DeepSHM applies a CNN over local sequence windows to predict mutation probabilities, capturing hotspot motifs but with no global view of the sequence~\citep{tang2022deep}. Thrifty widens the context by convolving over 5-mer embeddings while avoiding the parameter blow-up of explicit k-mer models~\citep{sung2025thrifty}. DASM separates mutation from selection explicitly, pairing a fixed SHM model with a learned amino-acid selection model~\citep{matsen2026separating}; its selection signal, however, is inferred from observed antibody evolution rather than measured binding, so it proxies functional fitness rather than affinity directly.
Architecturally, these approaches rely on local sequence contexts or predefined mutation models. In contrast, iGC Stage~1 formulates SHM as a full-sequence mutation-prediction problem. Given a parent-child pair (PCP), a Transformer encoder processes the entire VH sequence and jointly predicts both the mutation position and the resulting amino-acid substitution through dedicated position and amino-acid prediction heads. By conditioning on the complete antibody context, iGC can capture long-range dependencies that are inaccessible to local motif-based approaches.

\paragraph{Affinity-Guided Antibody Optimization.}

While SHM governs how diversity is generated, affinity maturation ultimately selects variants that bind better. This has driven a large body of affinity-aware design work, including supervised affinity predictors~\citep{shan2022deep}, active-learning loops~\citep{gessner2024active}, language-model sampling~\citep{hie2024efficient}, and diffusion- and flow-based generation~\citep{luo2022antigen, yu2026high}. Benchmarks such as EPAM~\citep{johnson2025nucleotide} and AbBiBench~\citep{zhao2025abbibench} have highlighted complementary challenges for computational affinity maturation. EPAM shows that incorporating SHM-aware evolutionary information substantially improves affinity-maturation prediction, whereas AbBiBench emphasizes accurate evaluation of antibody binding in the context of the target antigen. Taken together, they underscore the importance of modeling both biologically realistic mutation processes and affinity-driven selection. Existing methods, however, typically emphasize one objective at the expense of the other.

Among existing approaches, CoSiNE and SHMReplay are conceptually the closest to iGC, but neither directly trains its mutation generator to optimize antigen binding. CoSiNE learns antibody evolution as a neural continuous-time Markov process; its Guided Gillespie procedure then queries an external binding oracle during sampling to favor higher-scoring transitions~\citep{lu2026conditionally}. SHMReplay similarly learns to replay observed affinity-maturation trajectories and uses affinity scores from the GCReplay landscape to select or rank generated sequences~\citep{dewitt2025replaying}. Thus, their underlying mutation proposals remain binding-blind, with antigen specificity introduced through external oracle-guided selection rather than learned within the mutation policy itself.
Unlike these approaches, iGC explicitly formalizes affinity maturation as a reinforcement-learning process, drawing a direct parallel to the biological engine: SHM-driven diversification plays the role of exploration, while affinity-based selection drives policy improvement. Its central optimization mechanism is a reward--tax meta-gradient objective rather than a hand-designed adaptive shaping rule: RewardNet learns how to transform measured affinity signals, TaxNet learns a sequence-dependent penalty from evolutionary prevalence and SHM accessibility, and both are trained through their effect on a differentiable virtual policy update.

\section{The iGC Framework}
\label{sec:igc}
iGC models affinity maturation as a policy over single-mutation actions. Given a parent VH sequence, the policy chooses an action $(i,a)$, where $i$ is an aligned residue position and $a$ is the replacement amino acid. Stage~1 learns this action distribution from unlabeled parent-child pairs and domain-adapts it to the target biological context. Stage~2 then fine-tunes the adapted mutation prior under affinity-derived rewards while a KL penalty keeps proposals biologically plausible.

\begin{figure}[!hbt]
    \centering
    \includegraphics[width=\linewidth]{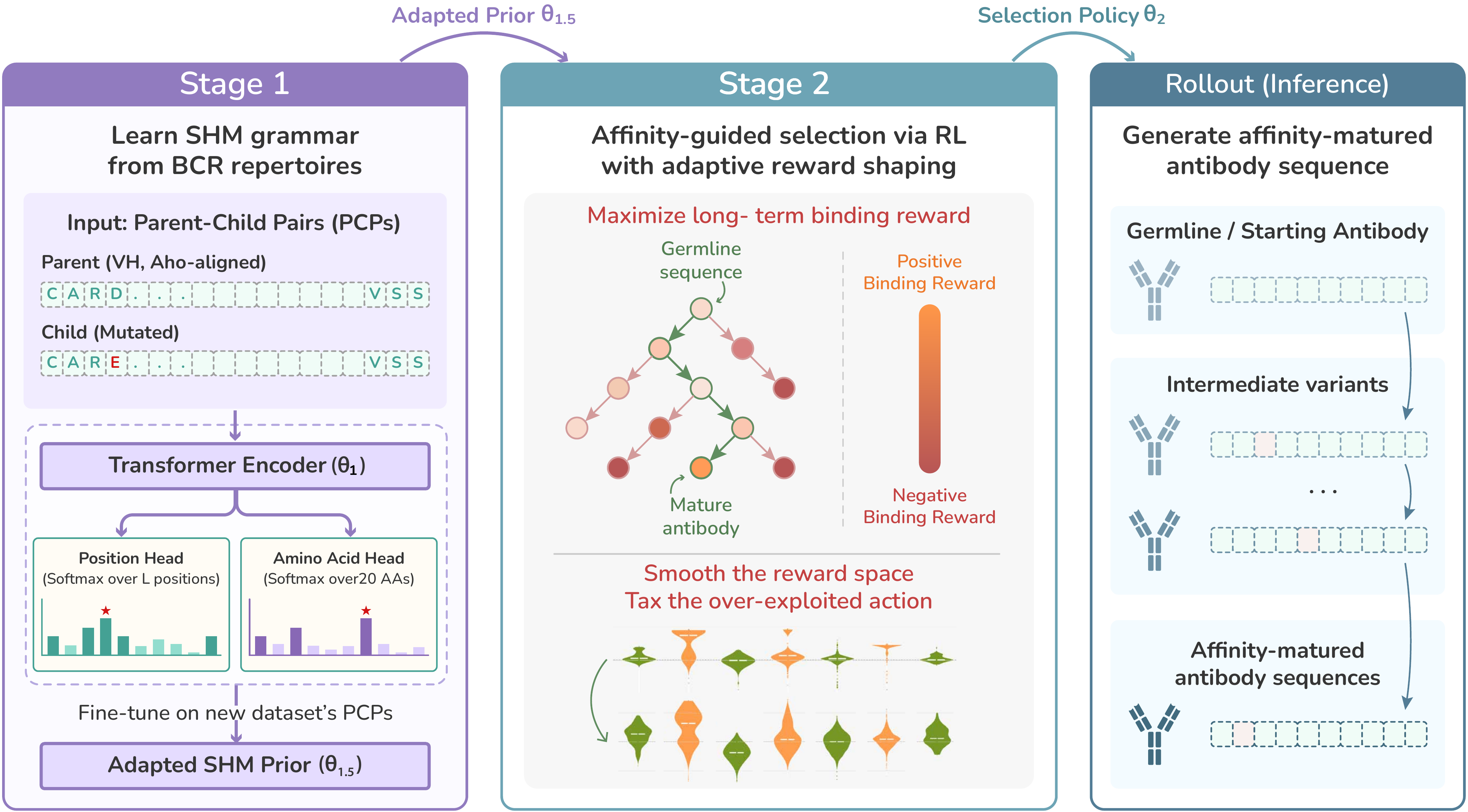}
    \caption{Overview of the two-stage iGC framework. Stage~1 learns and domain-adapts an SHM mutation prior, and Stage~2 optimizes the resulting policy for antigen binding while remaining regularized toward that prior. At inference, the learned policy optimizes an input antibody by rolling out successive mutations under a fixed mutation budget.}
    \label{fig:framework_pipeline}
\end{figure}

\subsection{Stage 1: Learning and Adapting SHM Grammar}
\label{subsec:igc_shm}
Stage~1 performs behavior cloning from observed mutation events, estimating a conditional prior $p(i,a\mid\mathbf{s})$ over one-step mutation actions. We represent each observed SHM event as a Parent-Child Pair (PCP): a parent antibody sequence, the mutated position under the Aho numbering scheme, and the resulting amino acid identity. Aho numbering is a standard antibody alignment convention that maps homologous residues from antibodies of different lengths onto shared position indices, making mutation positions comparable across sequences~\citep{honegger2001yet}. Each observed mutation contributes one PCP training example, allowing the model to learn from independent mutation events without requiring full lineage trajectories, fitness measurements, or affinity labels. This makes SHM data abundant and broadly available, allowing the model to learn mutation grammar purely from the natural distribution of SHM events.

Given a parent sequence $\mathbf{s} \in \mathcal{A}^L$ where $\mathcal{A}$ is the set of 20 standard amino acids and $L$ is the sequence length, a Transformer encoder processes $\mathbf{s}$ to produce contextual residue representations. The model learns the joint conditional distribution:
$ p(i, a \mid \mathbf{s}) = p(i \mid \mathbf{s}) \cdot p(a \mid i, \mathbf{s})$,
where $i \in \{1, \ldots, L\}$ is the mutation site and $a \in \mathcal{A}$ is the replacement amino acid. This factorization reflects the biological structure of SHM: site selection and substitution identity are coupled, as local sequence context around position $i$ influences which amino acid is favored. During training, teacher forcing is applied: the ground-truth position $i^{(n)}$ is used to condition the amino acid head, and the model is trained by minimizing:

\begin{equation}
    \mathcal{L} = -\frac{1}{N} \sum_{n=1}^{N} \left[ \log p(i^{(n)} \mid \mathbf{s}^{(n)}) + \log p(a^{(n)} \mid i^{(n)}, \mathbf{s}^{(n)}) \right]
\end{equation}

over a dataset of $N$ PCPs $\{(\mathbf{s}^{(n)}, i^{(n)}, a^{(n)})\}_{n=1}^{N}$. Stage~1 then domain-adapts this model with behavior cloning on parent--child pairs from the target affinity-maturation dataset. The resulting Adapted SHM Prior serves as the initialization and frozen KL reference for Stage~2, preserving the learned mutation grammar while correcting dataset-specific sequence and species mismatch.

\subsection{Stage 2: Affinity-Guided Clonal Selection}
\label{subsec:igc_sel}
Stage~2 turns the Adapted SHM Prior produced by Stage~1 into a policy that favors affinity-improving mutations while remaining close to realistic SHM behavior. Training is offline: the policy learns from logged mutation outcomes rather than interacting with a live experimental system. At evaluation time, the learned policy is rolled out \textit{in silico} to generate new antibody variants under a fixed mutation budget (Figure~\ref{fig:framework_pipeline}).

\paragraph{Affinity Maturation Data.}

Affinity supervision is provided by \textit{clonal trajectory data}, which records ordered mutation paths as sequences of parent, action, child, and affinity-change tuples. Each trajectory defines a sequential decision-making problem: the current antibody sequence is the state, a mutation is the action, and antigen-binding affinity provides the reward. The RL policy learns from these trajectories to maximize long-term reward, corresponding to the binding affinity of the final sequence produced after multiple mutation decisions. Our main experiments use GCReplay~\citep{dewitt2025replaying}. Dataset construction and implementation details are given in Appendices~\ref{app:data_protocols} and~\ref{sec:impl}, respectively.

\paragraph{Reinforcement Learning Formulation.}

For a trajectory $\tau$, the state $\mathbf{s}_{\tau,t}$ denotes the antibody sequence after $t$ mutations, and the action $u_{\tau,t}=(i_{\tau,t},a_{\tau,t})$ specifies a mutation site $i_{\tau,t}$ and its replacement amino acid $a_{\tau,t}$, producing the next sequence $\mathbf{s}_{\tau,t+1}$. The immediate reward is obtained from the dataset-provided \texttt{reward\_bind} value associated with the selected substitution. Specifically, for $(i_{\tau,t},a_{\tau,t})$, we use the DMS-measured change in binding to CGG, where positive values indicate improved binding and negative values indicate reduced binding: $r_{\tau,t}=\texttt{reward\_bind}(i_{\tau,t},a_{\tau,t})$. Because binding measurements are available only for single mutants, the binding affinity of a multi-mutant child is not directly evaluated. Instead, the return of a trajectory is defined as the cumulative reward of its constituent single mutations, $R_{\tau,t}=\sum_{k=t}^{T-1}r_{\tau,k}$. Consequently, the learning objective favors mutation trajectories whose constituent substitutions have larger cumulative DMS-measured binding improvements. We optimize the mutation policy $\pi_\theta(u\mid\mathbf{s})$ using a REINFORCE-style logged-action objective~\citep{williams1992simple}, with returns computed from the observed trajectories.


\paragraph{Raw-Reward Policy Collapse.}

\begin{figure}[t]
    \centering
    \includegraphics[width=\linewidth]{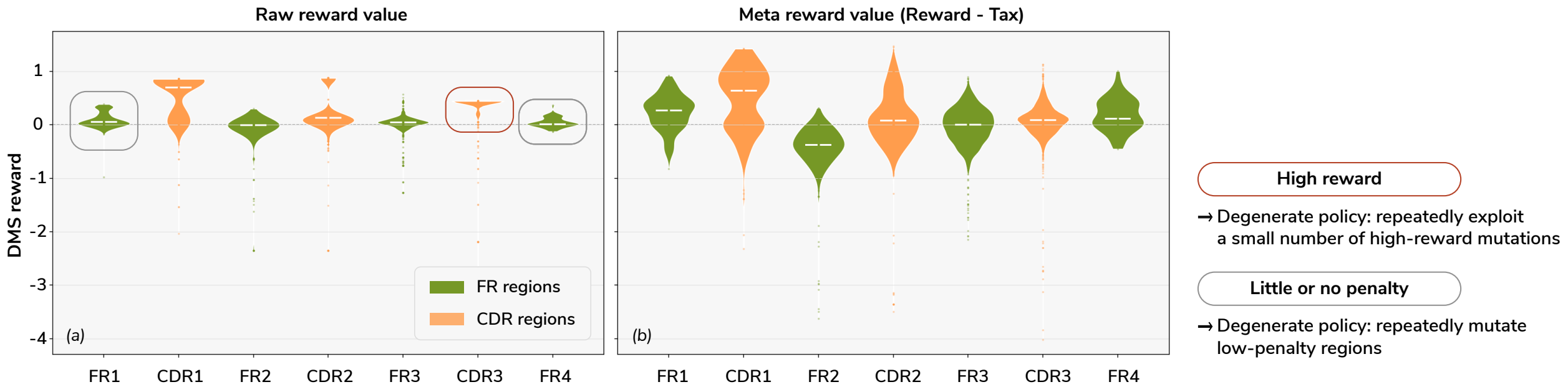}
    \caption{Reward landscape across antibody structural regions. \textit{(a)} Raw GCReplay rewards are imbalanced across regions, driving hotspot and safe-region policy collapse. \textit{(b)} The learned RTMG meta-reward smooths extreme peaks and produces a more balanced reward landscape.}
    \label{fig:reward_landscape}
\end{figure}

Directly optimizing the raw DMS reward is unstable because the reward distribution is highly skewed and varies substantially across antibody regions. A few large positive rewards can dominate the policy gradient, while low-variance framework mutations provide an easier but less informative optimization path. In GCReplay (Figure~\ref{fig:reward_landscape}a), these effects produce two failure modes: \textit{hotspot collapse}, in which the policy repeatedly selects a small set of high-reward mutations, and \textit{safe-region collapse}, in which it concentrates on structurally tolerant framework positions. Both modes improve the training objective without learning a diverse affinity-maturation policy. In contrast, the learned RTMG meta-reward smooths extreme reward peaks and reduces regional imbalance, producing the more even reward landscape shown in Figure~\ref{fig:reward_landscape}b.

\paragraph{SHM-Prior Initialization and Domain Adaptation.}

To keep affinity optimization aligned with biologically plausible SHM behavior, Stage~1 includes a supervised domain-adaptation phase on GCReplay parent--child pairs after pretraining on human repertoires. This phase corrects the human-to-mouse sequence mismatch and produces the Adapted SHM Prior. The trainable Stage~2 policy is initialized from this checkpoint, and a frozen copy, $\pi_{\mathrm{prior}}$, is retained as an SHM reference. A KL penalty toward $\pi_{\mathrm{prior}}$ constrains subsequent RL updates from departing too far from the learned SHM distribution. This prior controls \emph{where} the policy explores, but it does not correct the imbalance in \emph{what} the raw reward favors. RTMG addresses this remaining optimization problem.

\paragraph{Reward--Tax Meta-Gradient.}

RTMG adapts meta-gradient intrinsic-reward learning~\citep{zheng2018} to learn a training signal for antibody mutation policies. It introduces two auxiliary networks with parameters $\phi$ and $\psi$. RewardNet transforms the cumulative measured reward $c_{\tau,t}$ together with trajectory depth, mutation position, and antibody region into a benefit score. Its pre-normalized output is bounded to suppress extreme peaks, then centered so that its batch mean equals a fixed target without changing within-batch ordering. TaxNet receives the floored log score of the resulting sequence,
\begin{equation}
    \ell_{\tau,t}
    =\max\!\left\{
        \log P_{\mathrm{seq}}(\mathbf{s}_{\tau,t+1}),
        \ell_{\min}
    \right\},
\end{equation}
where $P_{\mathrm{seq}}$ combines empirical sequence frequency with SHM-prior accessibility (Appendix~\ref{app:sequence_probability}). Sequences that recur across convergent trajectories, or are readily reachable from recurrent sequences under the SHM prior, receive larger scores and can therefore incur a larger exploitation tax. For a training batch $\mathcal{B}$, the meta-reward is
\begin{equation}
    r_{\mathrm{meta}}^{(\tau,t)}
    =\overline{R}_{\phi,\mathcal{B}}^{(\tau,t)}
    -T_\psi(\ell_{\tau,t}),
\end{equation}
where $\overline{R}_{\phi,\mathcal{B}}^{(\tau,t)}$ is RewardNet's normalized benefit and $T_\psi(\ell_{\tau,t})$ is TaxNet's sequence-dependent penalty. RewardNet therefore determines \emph{which outcomes are valuable}, whereas TaxNet regulates \emph{how strongly recurrent or SHM-accessible sequence outcomes are exploited}.

RewardNet and TaxNet are not fitted to manually specified targets. Instead, they are optimized through the effect of $r_{\mathrm{meta}}$ on a temporary \emph{probe policy}. Let $\mathcal{L}_{\mathrm{inner}}(\theta;r_{\mathrm{meta}})$ denote the return-weighted logged-action REINFORCE objective plus the KL penalty toward $\pi_{\mathrm{prior}}$, as defined in Appendix~\ref{sec:impl}. One differentiable inner step gives
\begin{equation}
    \theta'
    =\theta-\eta_{\mathrm{in}}\nabla_\theta
        \mathcal{L}_{\mathrm{inner}}(\theta;r_{\mathrm{meta}}).
    \label{eq:probe-update}
\end{equation}
The parameters $\theta'$ are used only to evaluate the meta-objective and are never deployed. This objective encourages mutation-site diversity through Shannon entropy, capped at a target value so that diversity does not dominate once sufficient exploration is reached. Three Pearson-correlation losses preserve fidelity: i) RewardNet must remain aligned with cumulative DMS reward and with the log sequence-probability score of the trajectory's terminal sequence; ii) TaxNet must remain aligned with the per-step resulting-sequence score supplied to it; iii) and the net meta-reward must remain aligned with measured DMS reward. Differentiating this objective through Equation~\ref{eq:probe-update} updates only $(\phi,\psi)$. The entropy term determines whether the learned shaping improves exploration, while the fidelity terms prevent the auxiliary networks from inventing an arbitrary training signal.

After updating the auxiliary networks, we recompute the meta-reward and update the deployed policy with its gradient detached,
\begin{equation}
    \theta\leftarrow\theta-\eta_\theta\nabla_\theta
        \mathcal{L}_{\mathrm{inner}}
        \!\left(\theta;\operatorname{stopgrad}(r_{\mathrm{meta}})\right).
    \label{eq:policy-update}
\end{equation}
Thus, the second-order meta-gradient learns only the reward-shaping networks, whereas $\pi_\theta$ receives a standard first-order REINFORCE-plus-KL update. Both steps use the same logged trajectories; no mutation is sampled or experimentally executed during training. Algorithm~\ref{alg:rtmg} summarizes the update order, while Appendix~\ref{sec:impl} provides the architecture in Figure~\ref{fig:rtmg} and gives detailed definitions of the network outputs, normalization, returns, and outer objective.

\begin{algorithm}[t]
\caption{Reward--Tax Meta-Gradient training on logged trajectories}
\label{alg:rtmg}
\begin{algorithmic}[1]
\REQUIRE Logged trajectory data $\mathcal{D}$; Adapted SHM Prior $\pi_{\mathrm{prior}}$; policy $\pi_\theta$ initialized from $\pi_{\mathrm{prior}}$; RewardNet $R_\phi$; TaxNet $T_\psi$
\FOR{each training iteration}
    \STATE Sample a batch of logged transitions and sub-trajectories $\mathcal{B}\sim\mathcal{D}$.
    \STATE Compute RewardNet benefits $\overline{R}_{\phi,\mathcal{B}}$, sequence scores $\ell$, taxes $T_\psi(\ell)$, and meta-rewards $r_{\mathrm{meta}}=\overline{R}_{\phi,\mathcal{B}}-T_\psi(\ell)$.
    \STATE Compute Monte Carlo meta-reward returns and the inner REINFORCE-plus-KL loss.
    \STATE Form the differentiable probe update $\theta'$ using Equation~\ref{eq:probe-update}; do not overwrite $\theta$.
    \STATE Evaluate the capped-entropy term under $\pi_{\theta'}$ and the four fidelity losses for RewardNet, TaxNet, and their difference.
    \STATE Update $(\phi,\psi)$ by differentiating the outer objective through $\theta'$.
    \STATE Recompute $r_{\mathrm{meta}}$ with the updated auxiliary networks and detach it from $(\phi,\psi)$.
    \STATE Update the deployed policy $\theta$ using Equation~\ref{eq:policy-update}.
\ENDFOR
\ENSURE Affinity-optimized mutation policy $\pi_\theta$
\end{algorithmic}
\end{algorithm}

\section{Experimental Results}
\label{sec:results}
We first evaluate whether Stage~1 learns the mutation grammar of SHM, then test whether Stage~2 generates high-affinity, diverse antibody variants under fixed mutation budgets. Additional SHM probing of the optimized policy is reported in Appendix~\ref{app:final_igc_shm_probe}.

\subsection{Stage 1: SHM Mutation Prediction}
\label{subsec:res_shm}

\paragraph{Tang BCR Dataset.}
We evaluate the pretraining phase of Stage~1 on the Tang B-cell receptor (BCR) dataset~\citep{tang2022deep}, a BCR repertoire collection from 21 human donors processed through a standardized parent--child pair extraction pipeline~\citep{nguyen2015iq}, yielding 505,915 mutation pairs. Mutation patterns are consistent with known SHM biology: CDRs account for 45.4\% of all mutations despite spanning a minority of positions, and 84.2\% of substitutions are reachable through a single nucleotide change. The dataset is split at the donor level (70/15/15) to prevent repertoire leakage across splits. Full dataset statistics are provided in Appendix~\ref{app:tang}.

\begin{figure}[!hbt]
    \centering
    \includegraphics[width=\linewidth]{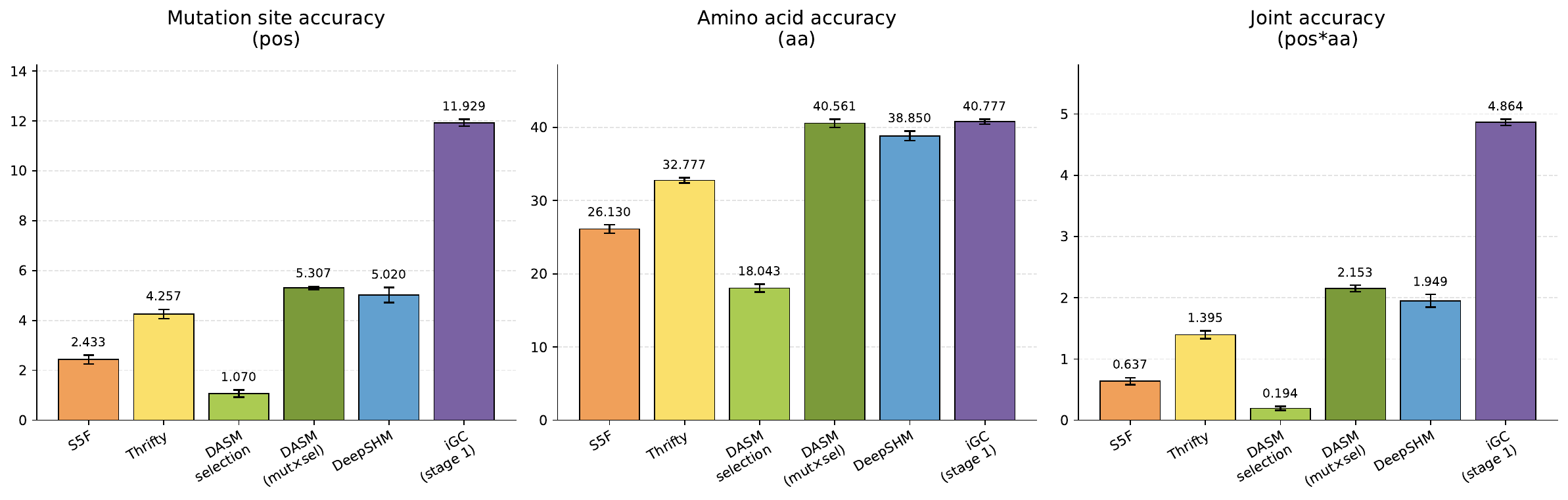}
    \caption{Stage~1 SHM mutation prediction on the Tang test set: top-1 mutation-site accuracy (left), top-1 amino-acid accuracy (middle), and joint accuracy (right) for each model. Bars show the mean over three data splits (in \%), with error bars indicating the standard error; values are the same as in Table~\ref{tab:shm_prediction_accuracy}.}
    \label{fig:stage1_benchmark}
\end{figure}

\paragraph{Benchmarking Setup.}
We evaluate Stage~1 against four SHM models: S5F~\citep{yaari2013models}, DASM (in two variants: mutation-and-selection and selection-only)~\citep{matsen2026separating}, Thrifty~\citep{sung2025thrifty}, and DeepSHM~\citep{tang2022deep}. The Stage~1 neural models are trained with batch size 256, and all models are evaluated on the Tang test set. SHM prediction is framed as two independent sub-tasks: given a parent sequence, predict (1) which position mutates next, evaluated by pos@$k$ (whether the true mutation site appears among the top-$k$ ranked positions), and (2) which amino acid it mutates to, evaluated by aa@$k$ (whether the observed substitution appears among the top-$k$ predicted amino acids at the true position). Conditioning the amino-acid sub-task on the ground-truth position isolates substitution accuracy from position accuracy, allowing fair comparison across models with fundamentally different architectures.
\paragraph{Results.}
Table~\ref{tab:shm_prediction_accuracy} reports top-$k$ position, amino-acid, and joint accuracy across all models, and Figure~\ref{fig:stage1_benchmark} visualizes the top-1 and joint results. iGC Stage~1 leads on all position metrics and on joint accuracy. The decisive gap is on position: pos@1 of $11.929\%$ more than doubles the best baseline (DASM mutation$\times$selection, $5.307\%$). Substitution accuracy is close to the strongest baselines (aa@1 $40.777\%$ versus DASM $40.561\%$), so the compounding of position and substitution performance gives a joint accuracy of $4.864\%$, more than twice the next best ($2.153\%$). The advantage is concentrated on \emph{where} rather than \emph{what}, suggesting that the transformer's full-sequence context captures hotspot targeting that motif-based and shallow learned models miss. Appendix~\ref{app:stage1_attention} further examines where the model attends when predicting mutation sites.

\begin{table*}[!hbt]
\centering
\caption{Prediction accuracy of mutation position and amino acid identity for different SHM models. Joint accuracy is the fraction of examples where both the predicted mutation position and amino acid substitution are correct. Values are reported as mean $\pm$ standard error over three data splits. Arrows indicate the preferred direction for each metric.}
\setlength{\tabcolsep}{2pt}
\resizebox{\linewidth}{!}{%
\begin{tabular}{lccccccc}
\toprule
\textbf{Model} & \textbf{pos@1 ($\uparrow$)} & \textbf{pos@3 ($\uparrow$)} & \textbf{pos@5 ($\uparrow$)} & \textbf{aa@1 ($\uparrow$)} & \textbf{aa@3 ($\uparrow$)} & \textbf{aa@5 ($\uparrow$)} & \textbf{Joint ($\uparrow$)} \\
\midrule
S5F & 2.433 $\pm$ 0.172 & 7.813 $\pm$ 0.293 & 12.237 $\pm$ 0.457 & 26.130 $\pm$ 0.569 & 64.310 $\pm$ 0.601 & 84.953 $\pm$ 0.331 & 0.637 $\pm$ 0.057 \\
Thrifty & 4.257 $\pm$ 0.183 & 10.633 $\pm$ 0.249 & 15.533 $\pm$ 0.230 & 32.777 $\pm$ 0.367 & 68.747 $\pm$ 0.450 & 86.813 $\pm$ 0.052 & 1.395 $\pm$ 0.066 \\
DASM selection-only & 1.070 $\pm$ 0.149 & 3.960 $\pm$ 0.163 & 6.947 $\pm$ 0.169 & 18.043 $\pm$ 0.548 & 38.497 $\pm$ 0.649 & 52.203 $\pm$ 0.790 & 0.194 $\pm$ 0.032 \\
DASM (mutation $\times$ selection) & 5.307 $\pm$ 0.058 & 13.133 $\pm$ 0.393 & 19.197 $\pm$ 0.382 & 40.561 $\pm$ 0.555 & \textbf{75.737 $\pm$ 0.365} & 89.435 $\pm$ 0.225 & 2.153 $\pm$ 0.052 \\
DeepSHM & 5.020 $\pm$ 0.302 & 11.747 $\pm$ 0.253 & 17.467 $\pm$ 0.299 & 38.850 $\pm$ 0.653 & 74.250 $\pm$ 0.644 & 88.470 $\pm$ 0.122 & 1.949 $\pm$ 0.105 \\
\textbf{iGC (Stage 1 pretraining)} & \textbf{11.929 $\pm$ 0.141} & \textbf{25.169 $\pm$ 0.217} & \textbf{33.836 $\pm$ 0.238} & \textbf{40.777 $\pm$ 0.323} & 75.552 $\pm$ 0.321 & \textbf{90.027 $\pm$ 0.188} & \textbf{4.864 $\pm$ 0.052} \\
\bottomrule
\end{tabular}
}
\label{tab:shm_prediction_accuracy}
\end{table*}

\subsection{Stage 2: Affinity-Guided Clonal Selection}
\label{subsec:res_sel}

\paragraph{GCReplay Dataset.}
We fine-tune Stage~2 on GCReplay~\citep{dewitt2025replaying}, a controlled in vivo affinity-maturation dataset of clonal trajectories from CGG-immunized mice. All sequences share the IGHV1-2 V-gene background, and each observed mutation step is paired with a DMS-derived binding reward ($\Delta K_D$ against CGG), providing sequential supervision for policy optimization. Cohort statistics, trajectory extraction, data splits, and sub-trajectory augmentation are detailed in Appendix~\ref{app:gcreplay}.
\textbf{Benchmarking Setup.}
We benchmark iGC against two external baselines, SHMReplay~\citep{dewitt2025replaying} and CoSiNE~\citep{lu2026conditionally}, as well as the Adapted SHM Prior produced by Stage~1 and three internal RL ablations: RL with raw reward, adaptive reward shaping, and RewardNet only. All methods use identical data splits and generation budgets. All Stage~2 models are trained with batch size 64. Stage~2 checkpoints are selected on the validation set by maximizing $\overline{R}_{600}\times\text{joint@1}\times\#\mathrm{Dist}$. For each checkpoint, the current policy generates 100 validation rollouts at each mutation budget $k\in\{2,3,4,5,6,7\}$, giving 600 samples in total; $\overline{R}_{600}$ is their mean cumulative DMS reward, and \#Dist is the number of distinct mutation identities across these rollouts. At each mutation budget, every method generates the same number of candidate sequences. For each data split and budget, candidates are ranked by cumulative DMS reward, and we evaluate the top $N\in\{100,30,20,10\}$ candidates. Within each top-$N$ set, DMS is the mean reward and \#Dist is the number of distinct mutation identities. We also report $\mathrm{Prod}=\mathrm{DMS}\times\#\mathrm{Dist}$, which measures whether a method achieves high reward without collapsing onto a narrow mutation repertoire. Table~\ref{tab:stage2_benchmark_mean} averages each metric equally across the six mutation budgets and reports means and standard deviations over three data splits. Full baseline configurations are described in Appendix~\ref{app:baselines}.

\begin{table*}[!hbt]
  \centering
  \caption{Stage~2 benchmarking results averaged equally over mutation budgets $k=2$ to $7$, with the same number of generated sequences at each budget, and reported over three data splits. Each model's mean row is followed by a parenthesized standard-deviation row across splits. DMS is mean reward, \#Dist is the number of distinct mutation identities, and $\mathrm{Prod}=\mathrm{DMS}\times\#\mathrm{Dist}$. Higher is better for all metrics. Bold and underlined Prod values denote the best and second-best means, respectively. $^{*}$ denotes RL-based methods initialized from the Adapted SHM Prior.}
  \label{tab:stage2_benchmark_mean}
  \setlength{\tabcolsep}{3pt}
  \resizebox{\linewidth}{!}{
  \begin{tabular}{lcccccccccccc}
    \toprule
    & \multicolumn{3}{c}{Top 100} & \multicolumn{3}{c}{Top 30} & \multicolumn{3}{c}{Top 20} & \multicolumn{3}{c}{Top 10} \\
    \cmidrule(lr){2-4}\cmidrule(lr){5-7}\cmidrule(lr){8-10}\cmidrule(lr){11-13}
    Model & DMS & \#Dist & Prod & DMS & \#Dist & Prod & DMS & \#Dist & Prod & DMS & \#Dist & Prod \\
    \midrule
    SHMReplay & $0.514$ & 257.6 & \underline{132.4} & $0.878$ & 99.3 & \underline{87.2} & $0.990$ & 70.7 & 70.0 & $1.164$ & 38.5 & \underline{44.8} \\
     & \stddev{0.018} & \stddev{2.3} & \stddev{5.3} & \stddev{0.033} & \stddev{1.5} & \stddev{3.4} & \stddev{0.040} & \stddev{0.7} & \stddev{3.1} & \stddev{0.047} & \stddev{0.4} & \stddev{1.7} \\
    \cmidrule(lr){1-13}
    CoSiNE & $0.019$ & 195.3 & 3.7 & $0.990$ & 70.4 & 69.7 & $1.155$ & 50.0 & 57.8 & $1.372$ & 27.5 & 37.7 \\
     & \stddev{0.078} & \stddev{17.0} & \stddev{15.7} & \stddev{0.073} & \stddev{2.5} & \stddev{5.6} & \stddev{0.077} & \stddev{2.3} & \stddev{4.5} & \stddev{0.058} & \stddev{1.6} & \stddev{2.7} \\
    \cmidrule(lr){1-13}
    Adapted SHM Prior & $0.185$ & 166.9 & 30.9 & $1.045$ & 62.8 & 65.6 & $1.195$ & 44.4 & 53.1 & $1.414$ & 25.0 & 35.4 \\
     & \stddev{0.121} & \stddev{5.4} & \stddev{19.1} & \stddev{0.143} & \stddev{5.7} & \stddev{5.0} & \stddev{0.160} & \stddev{5.2} & \stddev{4.1} & \stddev{0.154} & \stddev{3.1} & \stddev{2.0} \\
    \cmidrule(lr){1-13}
    RL w/ raw reward$^{*}$ & $0.844$ & 92.6 & 78.2 & $1.670$ & 36.0 & 60.1 & $1.800$ & 26.7 & 48.1 & $1.970$ & 16.3 & 32.1 \\
     & \stddev{0.298} & \stddev{7.4} & \stddev{30.9} & \stddev{0.152} & \stddev{0.9} & \stddev{4.8} & \stddev{0.158} & \stddev{1.3} & \stddev{4.7} & \stddev{0.152} & \stddev{1.0} & \stddev{3.6} \\
    \cmidrule(lr){1-13}
    Adaptive reward shaping$^{*}$ & $0.945$ & 108.2 & 102.2 & $1.795$ & 39.0 & 70.0 & $1.921$ & 28.4 & 54.6 & $2.070$ & 16.9 & 35.0 \\
     & \stddev{0.167} & \stddev{7.9} & \stddev{10.6} & \stddev{0.067} & \stddev{3.9} & \stddev{6.0} & \stddev{0.035} & \stddev{2.5} & \stddev{4.3} & \stddev{0.017} & \stddev{1.0} & \stddev{1.8} \\
    \cmidrule(lr){1-13}
    RewardNet only$^{*}$ & $0.344$ & 93.4 & 32.1 & $1.199$ & 46.4 & 55.6 & $1.353$ & 35.2 & 47.6 & $1.557$ & 22.2 & 34.6 \\
     & \stddev{0.150} & \stddev{19.9} & \stddev{14.0} & \stddev{0.191} & \stddev{9.2} & \stddev{5.4} & \stddev{0.194} & \stddev{7.0} & \stddev{4.3} & \stddev{0.209} & \stddev{3.7} & \stddev{2.5} \\
    \cmidrule(lr){1-13}
    RTMG w/o entropy$^{*}$ & $0.901$ & 90 & 81.09 & $1.695$ & 50 & 84.75 & $1.798$ & 42 & \textbf{75.52} & $1.946$ & 23 & 44.76 \\
     & \stddev{0.214} & \stddev{11.2} & \stddev{14.8} & \stddev{0.118} & \stddev{7.6} & \stddev{11.7} & \stddev{0.109} & \stddev{6.1} & \stddev{10.4} & \stddev{0.097} & \stddev{4.3} & \stddev{7.1} \\
    \cmidrule(lr){1-13}
    \textbf{iGC}$^{*}$ & $1.010$ & 149.4 & \textbf{150.9} & $1.663$ & 58.2 & \textbf{96.8} & $1.791$ & 41.7 & \underline{74.7} & $1.974$ & 23.5 & \textbf{46.4} \\
     & \stddev{0.201} & \stddev{13.0} & \stddev{16.1} & \stddev{0.157} & \stddev{10.5} & \stddev{10.5} & \stddev{0.157} & \stddev{6.2} & \stddev{5.5} & \stddev{0.137} & \stddev{2.1} & \stddev{1.3} \\
    \bottomrule
  \end{tabular}
  }
\end{table*}

\paragraph{Results.}
Table~\ref{tab:stage2_benchmark_mean} shows the central reward--diversity trade-off. The RL-based methods generally achieve higher DMS than the non-RL generators (SHMReplay, CoSiNE, and the Adapted SHM Prior), showing the benefit of affinity-guided policy optimization. However, direct reward optimization concentrates probability on a smaller set of beneficial actions: RL with raw reward has the lowest \#Dist at every cutoff, and adaptive reward shaping attains high DMS while still sacrificing substantial diversity. RewardNet alone does not resolve this trade-off. Full iGC achieves the highest Prod for the Top 100, Top 30, and Top 10 candidate sets and ranks second for Top 20, where RTMG without entropy is marginally higher. This ablation indicates that the entropy term generally improves the reward--diversity balance across the generated candidate pool, although the no-entropy variant remains competitive at one cutoff. Comparisons with RewardNet only still isolate the contribution of TaxNet: RewardNet refines the affinity signal, while the learned tax prevents the resulting policy from collapsing onto a narrow set of mutations. Affinity optimization retains measurable SHM structure, with predicted-position mutability remaining correlated with S5F (Table~\ref{tab:shm_probe_igc}; Appendix~\ref{app:final_igc_shm_probe}).


\section{Conclusion}

iGC models antibody affinity maturation by separating mutation generation from antigen-specific selection and then coupling the two processes through reinforcement learning. The Stage~1 Transformer learns an SHM prior from B-cell lineage data and substantially improves mutation-site prediction, while Stage~2 optimizes binding without discarding the sequence structure encoded by that prior. Reward--Tax Meta-Gradient shaping further prevents the policy from collapsing onto a small number of dominant substitutions or structurally tolerated framework sites. Across mutation-policy and structure-based evaluations, iGC achieves a favorable balance between predicted binding improvement, mutation diversity, and SHM-consistent behavior, supporting learned mutation--selection policies as a framework for iterative antibody optimization. Our evaluation is limited by its reliance on DMS-derived training rewards rather than direct experimental measurements of generated candidates.

\clearpage
\bibliographystyle{abbrvnat}
\bibliography{references}
\clearpage
\appendix

\section{Additional Stage 2 Results}
\label{app:additional_stage2_results}

\subsection{SHM Probing of the Final iGC Policy}
\label{app:final_igc_shm_probe}

Section~\ref{subsec:res_sel} establishes that iGC favors high-binding-reward mutations. The complementary question is whether the final RL-tuned policy still proposes mutations consistent with somatic hypermutation rather than drifting into unconstrained amino-acid optimization. We therefore compare canonical SHM motif frequencies at real mutation positions, iGC-predicted positions, and random background positions in Table~\ref{tab:shm_probe_igc}.

\begin{table}[!hbt]
  \centering
  \caption{Canonical SHM motif frequencies for the final iGC policy after Stage~2 fine-tuning.}
  \label{tab:shm_probe_igc}
  \setlength{\tabcolsep}{10pt}
  \begin{tabular}{lccc}
    \toprule
    & WRC & WA & GYW \\
    \midrule
    Real mutation positions & 26.9\% & 24.9\% & 18.7\% \\
    iGC predicted positions & 57.0\% & 1.0\% & 11.1\% \\
    Random background & 2.0\% & 19.4\% & 4.1\% \\
    \bottomrule
  \end{tabular}
\end{table}

iGC strongly enriches WRC and GYW motifs relative to random background, showing that affinity optimization preserves substantial AID-associated targeting structure. Its predicted-position mutability also remains correlated with the S5F model (Pearson $0.474$, Spearman $0.450$). WA motifs are underrepresented. Unlike the three-base WRC and GYW motifs, WA is a short two-base motif that already occurs at 19.4\% of random background positions; because many WA occurrences are not mutated, motif presence alone provides a less specific and noisier learning signal. The final policy may therefore capture these SHM mechanisms less readily. Together with the binding results in Table~\ref{tab:stage2_benchmark_mean}, these results support the two central properties of iGC: its mutations favor binding while remaining substantially aligned with SHM biology.

\subsection{Additional Stage 2 Validation}
\label{app:stage2_additional}

\subsubsection{Held-Out Sequence Rediscovery}
\label{app:rediscovery_analysis}

This analysis tests whether the improvement can be explained by memorization. We split GCReplay by germinal center, so all lineage trees in the test germinal centers are unseen during training. For each mutation budget from $k=2$ to $4$, iGC generates 5,000 variants and is evaluated by exact matches to naturally observed positive-reward sequences in these held-out trees. As shown in Table~\ref{tab:rediscovery_rate}, iGC rediscovers 32.0\%, 10.9\%, and 5.9\% of the eligible test sequences at $k=2,3,4$, respectively.

\begin{table}[!hbt]
  \centering
  \caption{Rediscovery of unseen positive-reward sequences in the held-out test set.}
  \label{tab:rediscovery_rate}
  \setlength{\tabcolsep}{8pt}
  \begin{tabular}{cccc}
    \toprule
    $k$ & Unseen positives rediscovered & Total unseen positives & Rediscovery rate \\
    \midrule
    2 & 54 & 169 & \textbf{32.0\%} \\
    3 & 41 & 375 & \textbf{10.9\%} \\
    4 & 15 & 255 & \textbf{5.9\%} \\
    \bottomrule
  \end{tabular}
\end{table}


\subsection{Critical Stage 2 Design Ablations}
\label{app:stage2_design_ablations}

We ablate two implementation choices that directly control the learned reward-shaping signal: the information provided to RewardNet and the lower bound applied to TaxNet's log sequence-probability score.

\subsubsection{RewardNet Input Selection}

Table~\ref{tab:rewardnet_input_selection} compares candidate RewardNet inputs. The full configuration combines cumulative DMS reward, mutation count, position identity, and antibody-region identity, and yields the highest average DMS reward.

\begin{table}[!hbt]
  \centering
  \caption{RewardNet input selection. The chosen configuration is shown in bold.}
  \label{tab:rewardnet_input_selection}
  \setlength{\tabcolsep}{5pt}
  \resizebox{\linewidth}{!}{
  \begin{tabular}{lcc}
    \toprule
    RewardNet input & Avg. DMS reward ($\uparrow$) & Avg. \#distinct mut. ($\uparrow$) \\
    \midrule
    \texttt{cum\_dms + pos\_onehot} & 0.526 & 129.3 \\
    \texttt{cum\_dms + mut\_count} & 0.743 & 116.6 \\
    \textbf{\texttt{cum\_dms + mut\_count + pos\_onehot + region\_onehot}} & \textbf{1.134} & \textbf{108.0} \\
    \bottomrule
  \end{tabular}
  }
\end{table}

\subsubsection{TaxNet Sequence-Score Floor}

Table~\ref{tab:taxnet_floor_selection} reports the sweep used to select the lower bound on TaxNet's log sequence-probability input. We use $\ell_{\min}=-4.0$, which provides the selected reward--diversity trade-off: its average DMS reward is close to the maximum at $-4.5$ (1.134 vs. 1.142), while retaining slightly more distinct mutations (108.0 vs. 106.3).

\begin{table}[!hbt]
  \centering
  \caption{TaxNet log sequence-score floor selection over mutation budgets $k=1$ to $7$. The selected floor is shown in bold.}
  \label{tab:taxnet_floor_selection}
  \setlength{\tabcolsep}{10pt}
  \begin{tabular}{ccc}
    \toprule
    Floor & Avg. DMS reward ($\uparrow$) & Avg. \#distinct mut. ($\uparrow$) \\
    \midrule
    $-6.0$ & 0.528 & 155.9 \\
    $-5.5$ & 0.263 & 178.1 \\
    $-5.0$ & 0.411 & 129.4 \\
    $-4.5$ & 1.142 & 106.3 \\
    $\mathbf{-4.0}$ & \textbf{1.134} & \textbf{108.0} \\
    $-3.5$ & 0.880 & 127.1 \\
    $-3.0$ & 0.703 & 119.3 \\
    \bottomrule
  \end{tabular}
\end{table}

\subsection{Training Pipeline Ablation}

We isolate the contribution of the two-stage pipeline by comparing three variants: \textbf{P1} omits Stage~1 and trains Stage~2 directly from random initialization; \textbf{P2} uses only the Tang-pretraining phase of Stage~1 before Stage~2; and \textbf{P3} uses the complete Stage~1, consisting of Tang pretraining followed by GCReplay domain adaptation, before Stage~2. All three use identical Stage~2 hyperparameters and reward shaping. Results are in Table~\ref{tab:pipeline_ablation}.

\paragraph{Is Stage 1 pretraining necessary? (P1 vs. P2).}

Tang pretraining provides an 18\% improvement in joint accuracy (0.146 vs. 0.124 on test) and a 23\% gain in aa@1, confirming that behavior cloning on human repertoire data provides a meaningful prior for amino-acid substitution grammar. Pretraining also accelerates convergence: under the validation $\overline{R}_{600}\times\text{joint@1}\times\#\mathrm{Dist}$ criterion, P2 selects epoch 4 vs. epoch 12 for P1. P1 achieves higher proposed mutation reward (+0.512 vs. +0.455); without a BC prior anchoring it to any positional bias, RL explores more freely and converges on high-reward positions. However, this comes at the cost of a spurious FR1 artifact (18.3\% of argmax predictions vs. 4.5\% empirical), analogous to the FR4 artifact in P2.

\paragraph{Does Tang pretraining transfer to GCReplay? (P2).}

The Tang-pretrained model shows substantial zero-shot transfer from human repertoires to mouse IGHV1-2 sequences before any RL fine-tuning. However, P2 carries a cross-species artifact: the human-trained positional prior concentrates 67.2\% of probability on FR4 (mouse J-gene is out-of-distribution), and despite the framework penalty, FR4 remains at 26.6\% after RL, 19$\times$ the empirical rate. CDR2, the second most mutated region empirically (26.7\%), is correspondingly under-targeted at 14.8\%.

\paragraph{Does domain adaptation resolve the OOD artifact? (P3 vs. P2).}

The domain-adaptation phase of Stage~1 directly addresses the cross-species mismatch by fine-tuning the Tang-pretrained prior on GCReplay trajectories before Stage~2. This reduces FR4 from 26.6\% to 1.2\% (near the empirical 1.4\%) and brings CDR2 targeting to 21.6\% (empirical: 26.7\%). Joint accuracy improves from 0.146 to 0.162 on test, while suppressing the FR4 artifact induced by the human-to-mouse J-gene distribution shift.

\begin{table}[t]
\centering
\caption{Pipeline ablation on GCReplay test set. FR4\% and CDR2\% are argmax region distributions; empirical rates are 1.4\% and 26.7\% respectively.}
\label{tab:pipeline_ablation}
\begin{tabular}{lccccc}
\toprule
\textbf{Pipeline} & \textbf{pos@1} & \textbf{aa@1} & \textbf{joint} & \textbf{FR4\%} & \textbf{CDR2\%} \\
\midrule
P1: No Stage~1 & 0.296 & 0.417 & 0.124 & 7.4\%  & 33.7\% \\
P2: Stage~1 pretrain only & 0.286 & 0.512 & 0.146 & 26.6\%$^\ddagger$ & 14.8\% \\
P3: Full Stage~1 & \textbf{0.301} & \textbf{0.539} & \textbf{0.162} & \textbf{1.2\%} & \textbf{21.6\%} \\
\bottomrule
\end{tabular}
\end{table}

{\small $^\ddagger$FR4 artifact from human-to-mouse J-gene distribution shift; see text.}

\paragraph{Cross-timepoint generalization (D20-only training).}

As a supplementary check, we train Stage~2 on D20 GCReplay trajectories only and evaluate on D15 validation and test sets. D20-only training matches or marginally exceeds the full pipeline on all metrics (joint: 0.165 vs. 0.162, aa@1: 0.551 vs. 0.539), with nearly identical regional distributions. This suggests that D15 and D20 carry similar positional targeting signal, and that the learned policy is not simply memorizing timepoint-specific patterns.

\section{Additional Stage 1 Results}
\label{app:additional_stage1_results}

We complement the main-text Stage~1 mutation-prediction benchmark by inspecting the model's attention patterns, probing whether the Adapted SHM Prior retains canonical SHM motif preferences, and characterizing how Stage~1 accuracy scales with training-set size.

\subsection{Stage 1 Attention Probing}
\label{app:stage1_attention}

\begin{figure}[!hbt]
    \centering
    \includegraphics[width=\linewidth]{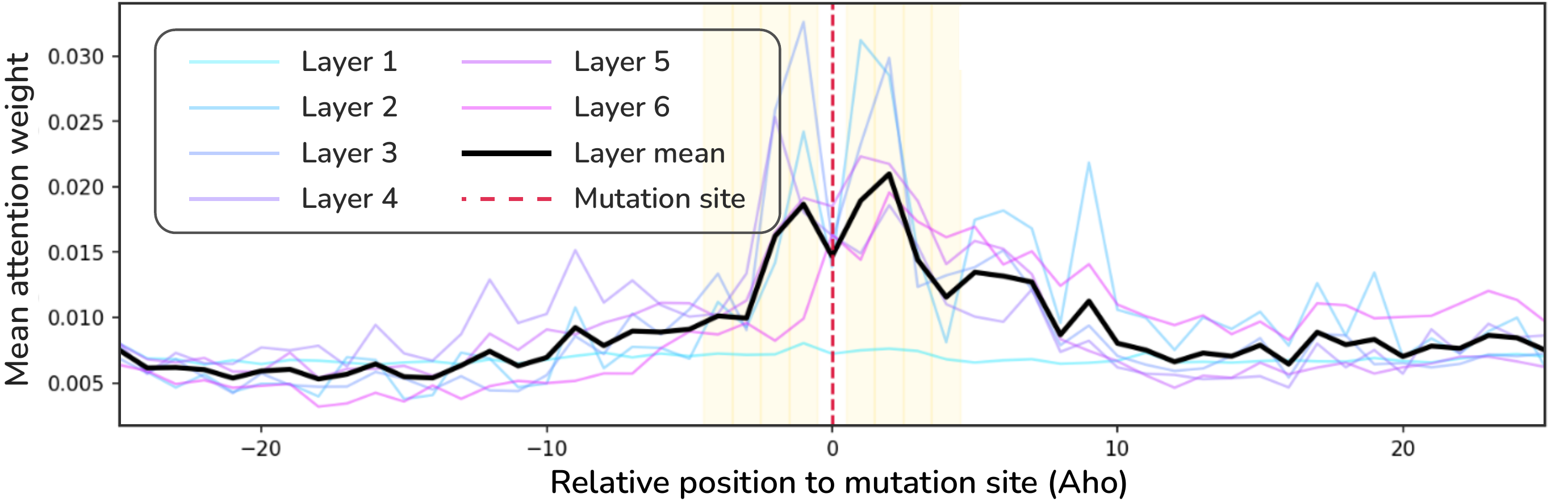}
    \caption{Mean attention weight per transformer layer as a function of relative sequence position to the predicted mutation site (position 0, dashed red), aggregated across the Tang BCR test set. Yellow bands highlight the $\pm$5 position window. All layers consistently attend most strongly to the local neighborhood of the mutation site, consistent with AID's known short-motif targeting mechanism.}
    \label{fig:attn_probe}
\end{figure}

Figure~\ref{fig:attn_probe} probes Stage~1 attention when predicting mutation sites. Mean attention peaks in the four- to five-position window around the true mutation site (highlighted in yellow), especially after averaging across layers. This local focus matches AID biology: cytosine deamination is driven by short sequence motifs (WRC/RCY), so mutation targeting depends strongly on nearby nucleotide context. The pattern suggests that the transformer learns SHM's local context dependence from observed mutation events, without explicit motif supervision.

\subsection{SHM-Prior Motif Probing}
\label{app:shm_probe_results}

Table~\ref{tab:shm_probe_prior} provides the corresponding diagnostic for the Adapted SHM Prior produced at the end of Stage~1, before Stage~2 fine-tuning. The Adapted SHM Prior has Pearson and Spearman correlations of 0.554 and 0.692 with S5F mutability, respectively.

\begin{table}[!hbt]
  \centering
  \caption{Canonical SHM motif frequencies for the Stage~1 Adapted SHM Prior.}
  \label{tab:shm_probe_prior}
  \setlength{\tabcolsep}{10pt}
  \begin{tabular}{lccc}
    \toprule
    & WRC & WA & GYW \\
    \midrule
    Real mutation positions & 26.9\% & 24.9\% & 18.7\% \\
    Adapted SHM Prior predicted positions & 54.4\% & 1.8\% & 42.1\% \\
    Random background & 2.0\% & 19.4\% & 4.1\% \\
    \bottomrule
  \end{tabular}
\end{table}

\subsection{Stage 1 Model Ablation}
\label{app:stage1_model_ablation}

We ablate the backbone used to learn the Stage~1 SHM mutation policy and track how this architectural choice propagates through the domain-adaptation phase of Stage~1 and subsequent Stage~2 fine-tuning.

\subsubsection{Backbone Architecture for SHM Policy Learning}

We ablate three backbone variants across the full two-stage pipeline, including both phases of Stage~1: a small from-scratch Transformer (4.8M parameters, $d=256$, 6 layers, 8 heads), a scaled from-scratch Transformer (14.3M, $d=384$, 8 layers, 12 heads), and AbLang2 (44.4M), an antibody-specific encoder pretrained on the Observed Antibody Space (OAS) via masked language modeling~\citep{olsen2022ablang}. AbLang2 uses rotary attention and SwiGLU activations; its MLM head is discarded and two linear heads for position and amino acid prediction are trained from scratch, while the encoder is fine-tuned end-to-end. The from-scratch and scaled models consume Aho-numbered alignments (fixed length 149); AbLang2 consumes raw heavy-chain sequences in its native tokenization, as its OAS-pretrained encoder implicitly encodes antibody sequence structure without explicit alignment.

We report joint accuracy as the fraction of examples for which both the mutation position and amino-acid identity are predicted correctly, along with the marginal top-1 position accuracy (pos@1) and amino-acid accuracy (aa@1). Results are summarized in Table~\ref{tab:backbone_ablation}.

\paragraph{Stage 1 pretraining and domain adaptation.}

During the Tang-pretraining phase of Stage~1, the final 4.8M from-scratch model trained on the full Tang dataset matches the main-text result: pos@1 = 11.929\%, aa@1 = 40.777\%, and joint@1 = 4.864\%. The scaled 14.3M model provides only a marginal joint-accuracy change (4.86\%), suggesting that capacity alone is not the limiting factor at this data scale. AbLang2 reaches a similar joint value in this ablation run (4.97\%) and converges faster, but the final pipeline uses the 4.8M from-scratch backbone for consistency with the main Stage~1 and Stage~2 experiments. After the GCReplay domain-adaptation phase of Stage~1, AbLang2 leads (15.87\% vs. 15.62\% for from-scratch), while the scaled model degrades noticeably (14.19\%), suggesting it overfits under distribution shift. These backbone comparisons are useful for model selection, but the reported full-Tang 4.8M result is kept consistent with Table~\ref{tab:shm_prediction_accuracy}.

\paragraph{RL fine-tuning (Stage 2).}

Under matched hyperparameters ($\text{lr} = 5\times10^{-5}$, $\beta_{\text{KL}} = 0.5$), the from-scratch model outperforms AbLang2 on joint after RL training (15.20\% vs. 13.37\%). The validation $\overline{R}_{600}\times\text{joint@1}\times\#\mathrm{Dist}$ criterion selects epoch 1 for both models, a characteristic pattern of REINFORCE training: since RL optimizes for reward rather than pattern matching, the joint metric, which measures alignment with empirical mutation patterns, degrades as training progresses. The from-scratch model's RL-stage advantage likely reflects that its smaller capacity makes it less susceptible to drifting from the BC prior under policy gradient updates, even under the same KL penalty weight. We note that joint accuracy is an imperfect proxy for RL-stage quality, as it does not directly measure binding improvement; downstream evaluation on a binding affinity oracle would better distinguish backbone quality at this stage.

\begin{table}[t]
\centering
\caption{Backbone ablation across training phases of the two-stage pipeline. Entries report joint@1 accuracy (\%).}
\label{tab:backbone_ablation}
\begin{tabular}{lccc}
\toprule
\textbf{Training phase} & \textbf{From-scratch} & \textbf{Scaled} & \textbf{AbLang2} \\
               & (4.8M) & (14.3M) & (44.4M) \\
\midrule
Stage 1: Tang pretraining & 4.864\% & 4.86\% & \textbf{4.97\%} \\
Stage 1: GCReplay adaptation & 15.62\% & 14.19\% & \textbf{15.87\%} \\
Stage 2: GCReplay RL & \textbf{15.20\%} & 9.47\% & 13.37\% \\
\bottomrule
\end{tabular}
\end{table}

\subsection{Stage 1 Data Scaling}
\label{app:stage1_scaling}

\begin{figure}[!hbt]
    \centering
    \includegraphics[width=\linewidth]{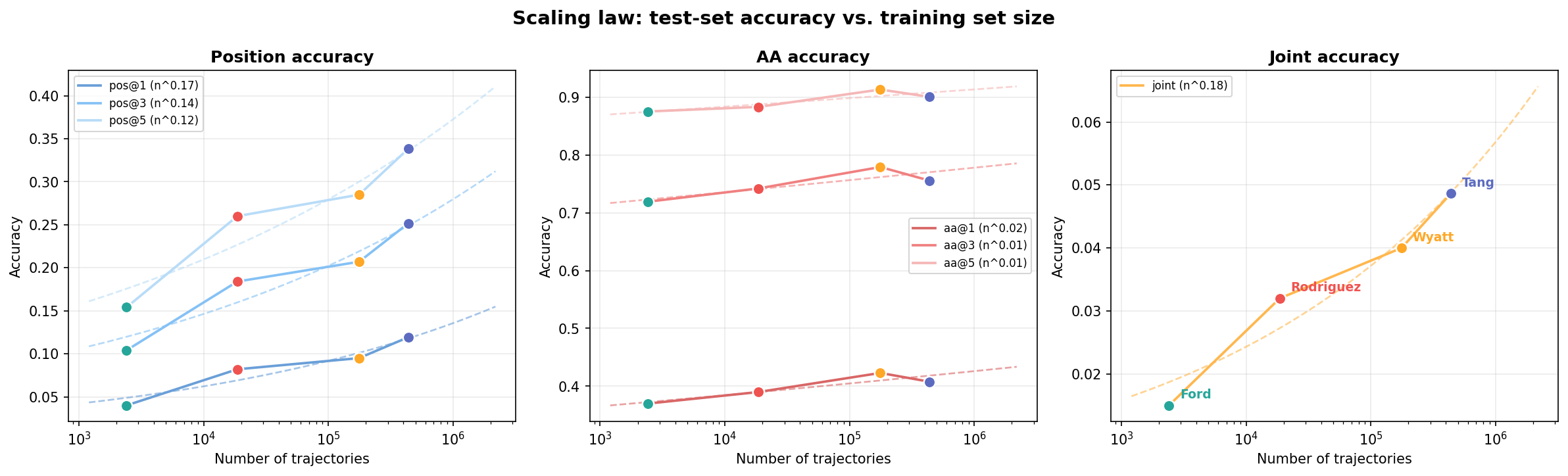}
    \caption{Stage~1 scaling law: test-set accuracy versus number of training trajectories for position (left), amino-acid (middle), and joint (right) prediction. Each point is a separate model trained on one PCP dataset (Ford, Rodriguez, Jaffe, Tang; the Jaffe point is labeled ``Wyatt'' in the right panel). Dashed lines are power-law fits $\text{acc} = a\,n^{b}$ extrapolated beyond the largest dataset; fitted exponents are given in the legends.}
    \label{fig:scaling_law}
\end{figure}

To understand how much of Stage~1 performance is limited by data, we train the same 4.8M-parameter Transformer policy (Appendix~\ref{app:stage1_model_ablation}) independently on four human heavy-chain parent--child-pair (PCP) collections of increasing size, compiled by~\citet{johnson2025nucleotide}: Ford~\citep{ford2023flairr} (2,407 trajectories), Rodriguez~\citep{rodriguez2023genetic} (18,639), Jaffe~\citep{jaffe2022functional} (176,656; labeled ``Wyatt'' in Figure~\ref{fig:scaling_law} after the archive's file prefix), and Tang~\citep{tang2022deep,vergani2017novel} (438,302). The Ford, Rodriguez, and Jaffe models are single runs with Aho-numbered input, batch size 512, learning rate $10^{-3}$ with 1,000 warmup steps, 10 epochs, and seed 42; the Tang point is the main-text Stage~1 result (Table~\ref{tab:shm_prediction_accuracy}; batch size 256, mean over three data splits). All models use donor-stratified 70/15/15 splits. We report top-$k$ position accuracy (pos@$k$), amino-acid accuracy at the true site (aa@$k$), and joint top-1 accuracy on each dataset's held-out test split, and fit power laws $\text{acc} = a\,n^{b}$ in the number of training trajectories $n$ (Figure~\ref{fig:scaling_law}).

\paragraph{Position prediction scales with data; amino-acid prediction saturates.}
Position accuracy improves steadily as the number of training trajectories grows, and the power-law fits show that it keeps rising with more data across all top-$k$ cutoffs. In contrast, amino-acid accuracy is nearly flat: even the smallest dataset already performs close to the largest, so adding data barely changes substitution prediction. This is consistent with the observation that most SHM substitutions are reachable by a single nucleotide change, so local codon context already supplies much of the amino-acid signal. Joint accuracy therefore inherits the position scaling, indicating that \emph{where} to mutate, rather than \emph{what} to mutate to, is the main data-limited component of Stage~1.

\section{Datasets}
\label{app:data_protocols}

We use one repertoire dataset to learn the SHM prior and two affinity-labeled datasets for reinforcement-learning experiments.

\subsection{BCR Tang for SHM Prior Learning}
\label{app:tang}
The Tang dataset is a paired B-cell receptor (BCR) repertoire collection from 21 human donors spanning three cohorts: ten healthy B-series donors (B10 to B21), seven chronic lymphocytic leukaemia (CLL) donors, and three healthy AG-series donors. Raw repertoires are processed through the Matsen lab parent-child pair (PCP) extraction pipeline, comprising IgBLAST V/J annotation (assignment of each sequence to its inherited variable and joining gene segments), DefineClones clonal clustering, IQ-TREE phylogenetic reconstruction, and HKY+G ancestral sequence reconstruction, producing directed mutation pairs where each edge in a clonal lineage tree yields one PCP. In total, the dataset contains 505,915 PCPs across 43,692 clonal lineage trees, decomposing into 438,302 trajectories and 2,024,413 individual single-amino-acid mutation steps. All sequences are IgH variable domains averaging 121 amino acids in length.

\paragraph{V and J Gene Usage.}

V gene usage spans 117 distinct IGHV genes, with IGHV3-23*01 the most frequent at 12.7\%. J gene usage spans 13 distinct IGHJ genes, with IGHJ4*02 the most frequent at 55.7\%. Inter-donor divergence in V-gene usage is substantial (mean pairwise Jensen-Shannon distance 0.487). CLL donors exhibit a more restricted, oligoclonal V-gene repertoire relative to healthy donors, consistent with the clonal expansion characteristic of the disease.

\paragraph{Mutation Statistics.}

Mutation burden per trajectory is heavily right-skewed: median 2, mean 4.62, max 47. Mutations concentrate in known SHM hotspots at IMGT positions 35 to 36 (CDR1) and 64 (CDR2). CDRs collectively account for 45.4\% of all mutations despite spanning a much smaller fraction of scored positions, consistent with activation-induced cytidine deaminase (AID) targeting of WRC/RCY motifs. 84.2\% of observed amino acid substitutions are reachable via a single nucleotide change from the parent codon, matching AID's single-cytosine-deamination mechanism. The remaining 15.8\% reflect multiple AID events or repair-patch mutations accumulated on the same tree edge, rather than sequencing artefacts.

\paragraph{Dataset Splits.}

The dataset is split at the donor level using greedy bin-packing targeting a 70/15/15 ratio (seed=42), ensuring no donor appears in more than one split. This yields 293,227 trajectories (12 donors) for training, 71,436 (4 donors) for validation, and 73,639 (5 donors) for test, corresponding to 1,439,804, 295,130, and 289,479 non-terminal mutation steps respectively, out of 2,024,413 total steps. V gene usage, J gene usage, and mutation burden distributions are statistically indistinguishable across the three splits, confirming that any generalisation gap reflects model capacity rather than distributional shift.

\subsection{GCReplay for RL Fine-Tuning}
\label{app:gcreplay}
GCReplay is a controlled in vivo affinity maturation experiment in which 25 mice are immunized with CGG (chicken gamma-globulin, the antigen target in this assay) and germinal-center B cells are sampled at two timepoints (day 15 and day 20 post-immunization) across 119 distinct germinal centers (4 to 5 per mouse, 14 to 77 trajectories per GC). Unlike the BCR Tang dataset, every sequence in GCReplay derives from a single fixed V-gene background (IGHV1-2), yielding a constant 112-amino-acid heavy-chain length. Every observed mutation step carries a directly measured fitness signal: \texttt{reward\_bind} ($\Delta K_D$ against CGG, derived from deep mutational scanning, where single mutants are experimentally scored for binding). We use two trajectory views of GCReplay. For descriptive edge-level statistics, each clonal-tree edge is decomposed into its own short mutation trajectory (6,188 edge trajectories and 23,128 edge-derived sub-trajectories; median 1 mutation per edge trajectory; 65\% single-amino-acid edges), preserving raw SHM events including detrimental mutations. For Stage~2 training and evaluation, we use the root-to-leaf trajectory view described below.

\paragraph{Mutation Statistics.}
71.8\% of mutations are beneficial (versus a 50\% null expectation), confirming strong positive selection. Mutational mass concentrates in FR3 (32.8\%, near-neutral reward) and CDR2 (26.7\%, the dominant antigen-contact loop for IGHV1-2), while CDR3 and CDR1 carry the largest per-mutation binding gains (mean $\Delta K_D$ of +0.291 and +0.259 respectively). 98\% of substitutions are single-nucleotide accessible, compared to 84\% in Tang, consistent with the tighter evolutionary constraint of a single in vivo experiment. Day-20 germinal centers show higher and less variable binding gains than day-15 germinal centers, consistent with progressive affinity maturation over time.

\paragraph{Trajectory Extraction.}

The Stage~2 policy uses a separate root-to-leaf trajectory view. Trajectories are extracted from each clonal tree and split by germinal center into training, validation, and test sets containing 1,445, 1,046, and 997 trajectories, respectively. These counts are not intended to match the edge-level counts of 6,188 and 23,128 reported above, which describe a different extraction stage. For RL training, the 1,445 training trajectories contain 12,048 action steps; enumerating all contiguous sub-trajectories expands them into 75,022 training items. Under this construction, a transition at position $k$ in a trajectory of length $n$ appears in $n-k+1$ training items, so earlier transitions receive greater weight than later transitions and the augmented state distribution differs from the original trajectory distribution. We do not importance-correct this augmentation-induced reweighting. The validation set remains at the full-trajectory level without augmentation. All sub-trajectories are retained regardless of reward sign, preserving detrimental mutations in the training distribution.

\paragraph{Domain Gap.}

GCReplay's single mouse IGHV1-2 background differs substantially from Tang's diverse human V-gene repertoire. The Aho-149 canonical alignment bridges this gap at the sequence level by placing residues from different antibodies into a shared 149-position coordinate system, and the domain-adaptation phase of Stage~1 continuously trains on GCReplay PCPs to adapt the SHM grammar prior to the murine sequence context before Stage~2 begins.

\section{Benchmarking Protocols}
\label{app:benchmark_protocols}

\subsection{Stage 1 Benchmark Protocol}

All Stage~1 models are evaluated on the held-out Tang donor split. We separately measure mutation-site ranking with pos@$k$, amino-acid prediction at the true site with aa@$k$, and joint top-1 accuracy, so improvements in site selection can be distinguished from improvements in substitution identity.

\subsection{Stage 2 Benchmark Protocol}

\paragraph{Baseline Configurations.}
\label{app:baselines}

\paragraph{SHMReplay.}

SHMReplay is run in its native fixed-k-mutation rollout mode (Strategy I), where mutations are applied unconditionally regardless of predicted DMS score and scored post hoc by the common structure-based proxy. No acceptance filter is applied during generation, meaning the method reflects the raw SHMReplay mutation distribution without any affinity-guided steering.

\paragraph{CoSiNE.}

CoSiNE uses an ESM2-150M encoder with a FreeRateMatrix continuous-time Markov chain (CTMC) head trained on GCReplay. Candidate sequences are generated by standard discrete Gillespie simulation directly from the learned CTMC substitution rates, without affinity-oracle reweighting or Guided Gillespie steering. This provides a fair comparison of the learned mutation policies rather than giving CoSiNE access to an additional oracle during generation. No position is revisited within a rollout.

\paragraph{Adaptive Reward Shaping.}

This baseline uses the hand-designed adaptive reward-shaping rule from the previous version of the method. It reshapes the per-step reward before each policy-gradient update using one penalty for each observed failure mode. For a mutation at position $i$ in region $\rho \in \{\mathrm{CDR1\!-\!3, FR1\!-\!4}\}$, let $r$ denote the affinity-derived reward for the current mutation, $\mu$ and $\sigma$ the mean and standard deviation of the training reward distribution over all mutations, and $\mu_\rho^+$ the mean positive reward within region $\rho$. The shaped reward is

\begin{equation}
\tilde{r}(i,\rho)=
r-
\begin{cases}
\alpha r, & r > \mu+\sigma
\quad\text{(hotspot penalty)},\\[2mm]
(1-\mu_\rho^+)\lambda, & \text{otherwise (region penalty)}.
\end{cases}
\end{equation}

Here, $\alpha$ specifies the fraction of the current reward subtracted from hotspot mutations, and $\lambda$ controls the strength of the region penalty.

Hotspots are identified from the empirical reward distribution: any mutation whose reward exceeds one standard deviation above the global mean ($\mu+\sigma$) is treated as a hotspot. Applying the hotspot penalty suppresses exceptionally rewarding mutations without eliminating their reward advantage, preventing the policy from collapsing onto a small number of dominant mutation events. For all remaining mutations, the region penalty scales with $(1-\mu_\rho^+)$, assigning larger penalties to regions whose positive mutations are less informative and smaller penalties to regions that consistently contain beneficial mutations. The two penalties are mutually exclusive, ensuring that each mutation is corrected only for its dominant failure mode.

Shaping is applied independently to every mutation before returns are computed. For a sub-trajectory of length $T$, the shaped return is
\[
G=\sum_{t=1}^{T}\tilde r_t,
\]
which weights the log-probabilities of the corresponding actions in the REINFORCE update, while the KL regularization term keeps the learned policy close to the SHM prior.
\paragraph{Shared Evaluation Protocol.}

All compared methods start from the same IGHV1-2 germline sequence (the inherited, pre-mutation antibody sequence for this V-gene background; Clone 2.1 Fab is reverted to germline with substitutions A28D, R49K, and G64S) and generate 100 candidate sequences at each CDR-mutation budget from $k = 2$ to $7$. We write substitutions as X$n$Y, where X is the parent amino acid, $n$ is the heavy-chain residue position, and Y is the replacement amino acid; site-only labels such as A106 omit Y. Each candidate is evaluated in three independent modeling runs, which are averaged, using the 9ODB crystal structure of Clone 2.1 Fab bound to chicken IgY CH2, a CGG surrogate antigen. The common structure-based pipeline computes $\Delta\Delta G_\text{bind}$ relative to the germline structure, with more negative values indicating stronger predicted binding.

\section{Implementation Details}
\label{sec:impl}

\subsection{Evolutionary Sequence-Probability Estimation}
\label{app:sequence_probability}

TaxNet uses a relative sequence-probability score that combines evolutionary prevalence in GCReplay with accessibility under the frozen SHM prior. We first extract the trajectory collection $\mathfrak{T}$ from all GCReplay lineage trees. Every sequence at every step is placed into a multiset,
\begin{equation}
    \mathcal{M}
    =\biguplus_{\tau\in\mathfrak{T},\,0\le t\le T_\tau}
      \{\mathbf{s}_{\tau,t}\},
\end{equation}
where the joint index ranges over every trajectory $\tau$ and every step $t$ within that trajectory, and $\biguplus$ denotes multiset union. We deliberately retain duplicate sequences. A sequence reached by multiple branches or trajectories therefore receives a larger count, preserving the signal of convergent evolution; such recurrence commonly indicates that the sequence has been repeatedly selected during affinity maturation. The germline sequence is also repeated because it begins every trajectory.

Let $n(\mathbf{x})$ be the multiplicity of sequence $\mathbf{x}$ in $\mathcal{M}$, let $\mathcal{S}_{\mathrm{obs}}$ be the set of unique observed sequences, and let $\mathbf{g}$ denote the germline. The empirical frequency and germline-normalized prevalence of an observed sequence are
\begin{equation}
    \widehat p_{\mathrm{pool}}(\mathbf{x})
    =\frac{n(\mathbf{x})}{|\mathcal{M}|},
    \qquad
    P_{\mathrm{emp}}(\mathbf{x})
    =\frac{\widehat p_{\mathrm{pool}}(\mathbf{x})}
           {\widehat p_{\mathrm{pool}}(\mathbf{g})}
    =\frac{n(\mathbf{x})}{n(\mathbf{g})}.
\end{equation}
Thus, $P_{\mathrm{emp}}(\mathbf{g})=1$. The score preserves relative recurrence while using the germline, which appears in all trajectories, as a common reference.

For an unseen sequence $\mathbf{y}\notin\mathcal{S}_{\mathrm{obs}}$, let $d_{\mathrm{edit}}$ denote sequence edit distance and define its nearest observed set as
\begin{equation}
    d_{\min}(\mathbf{y})
    =\min_{\mathbf{x}\in\mathcal{S}_{\mathrm{obs}}}
      d_{\mathrm{edit}}(\mathbf{x},\mathbf{y}),
    \qquad
    \mathcal{S}_{\min}(\mathbf{y})
    =\left\{
        \mathbf{x}\in\mathcal{S}_{\mathrm{obs}}:
        d_{\mathrm{edit}}(\mathbf{x},\mathbf{y})=d_{\min}(\mathbf{y})
      \right\}.
\end{equation}
Let $P_{\mathrm{prior}}(\mathbf{y}\mid\mathbf{x})$ be the transition probability assigned by the frozen Adapted SHM Prior produced by Stage~1 for producing $\mathbf{y}$ from $\mathbf{x}$. We extend the empirical score to arbitrary sequences by propagating probability mass from every nearest observed sequence,
\begin{equation}
    P_{\mathrm{seq}}(\mathbf{y})
    =
    \begin{cases}
        P_{\mathrm{emp}}(\mathbf{y}),
        & \mathbf{y}\in\mathcal{S}_{\mathrm{obs}},\\[2mm]
        \displaystyle
        \sum_{\mathbf{x}\in\mathcal{S}_{\min}(\mathbf{y})}
        P_{\mathrm{emp}}(\mathbf{x})
        P_{\mathrm{prior}}(\mathbf{y}\mid\mathbf{x}),
        & \mathbf{y}\notin\mathcal{S}_{\mathrm{obs}}.
    \end{cases}
    \label{eq:sequence_probability}
\end{equation}
Equation~\ref{eq:sequence_probability} assigns an unseen sequence a large score when it is both close to recurrent evolutionary states and likely under the SHM mutation prior. Because the scores are normalized relative to the germline rather than over the complete sequence space, $P_{\mathrm{seq}}$ is a relative probability score, not a probability mass function required to sum to one. RTMG uses only its logarithm and relative variation: TaxNet receives the score of each resulting sequence $\mathbf{s}_{\tau,t+1}$, while the RewardNet fidelity objective uses the score of the terminal sequence $\mathbf{s}_{\tau,T_\tau}$.

\subsection{Reward--Tax Meta-Gradient Details}

This section gives the mathematical specification and bilevel optimization procedure for Reward--Tax Meta-Gradient, complementing the overview in Section~\ref{subsec:igc_sel}.

\begin{figure}[!hbt]
    \centering
    \includegraphics[width=\linewidth]{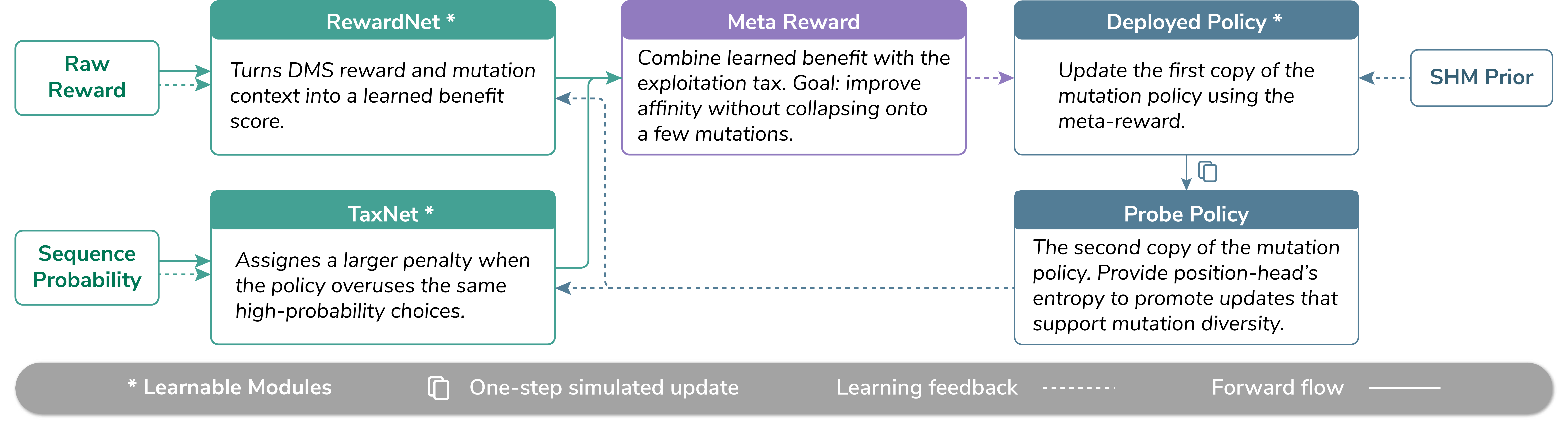}
    \caption{Reward--tax meta-gradient (RTMG) architecture. RewardNet converts the measured DMS reward and mutation context into a learned benefit, while TaxNet penalizes recurrent or SHM-accessible sequence outcomes according to their evolutionary sequence-probability score. Their difference defines the meta-reward used for a differentiable update of the probe policy; feedback from the probe policy then trains the reward and tax networks to improve affinity while preserving mutation diversity. The deployed policy is updated using the resulting meta-reward.}
    \label{fig:rtmg}
\end{figure}

\textbf{RewardNet} learns a refined reward from the measured DMS binding signal together with the mutation history. It takes as input the cumulative measured binding reward $c_{\tau,t}$, trajectory depth $d_{\tau,t}$, the mutated position represented as a one-hot vector $\mathbf{e}_{i_{\tau,t}}$, and the antibody region represented as a one-hot vector $\mathbf{e}_{\rho_{\tau,t}}$ over FR1--FR4 and CDR1--CDR3. Here, $\tau$ denotes the trajectory index, $t$ denotes the mutation step within the trajectory, $i_{\tau,t}$ denotes the mutated position index, and $\rho_{\tau,t}$ denotes its antibody region. It outputs a scalar reward that suppresses extreme reward peaks while retaining positive linear agreement with the affinity variation measured by DMS.

\begin{equation}
    R_\phi(c_{\tau,t},d_{\tau,t},\mathbf{e}_{i_{\tau,t}},\mathbf{e}_{\rho_{\tau,t}})
    =C_R\tanh\!\left(
        \frac{f_\phi([c_{\tau,t},d_{\tau,t},\mathbf{e}_{i_{\tau,t}},\mathbf{e}_{\rho_{\tau,t}}])}{C_R}
    \right),
\end{equation}
where $f_\phi$ is a multilayer perceptron. The Pearson-correlation terms used to train RewardNet preserve relative variation but do not determine the absolute scale or offset of its output. The $C_R\tanh(\cdot/C_R)$ parameterization bounds the pre-normalized reward $R_\phi$ to $[-C_R,C_R]$, preventing unbounded scale growth. Let $R_\phi^{(b)}$ denote this bounded output for sample $b$. We remove batch-dependent offset drift by pinning its batch mean to a fixed target $\mu_{\mathrm{reward}}$, a training hyperparameter that is independent of the current batch. For a training batch $\mathcal{B}$, we apply the exact forward-pass normalization
\begin{equation}
    \overline{R}_{\phi,\mathcal{B}}^{(b)}
    =R_\phi^{(b)}
    -\frac{1}{|\mathcal{B}|}\sum_{b'\in\mathcal{B}}R_\phi^{(b')}
    +\mu_{\mathrm{reward}},
\end{equation}
so that
\begin{equation}
    \frac{1}{|\mathcal{B}|}\sum_{b\in\mathcal{B}}\overline{R}_{\phi,\mathcal{B}}^{(b)}
    =\mu_{\mathrm{reward}}
\end{equation}
holds exactly for every training batch. This transformation adds the same constant to every sample after centering, so it preserves within-batch pairwise differences, ordering, and Pearson correlations. Because the mean shift is applied after the $\tanh$ bound, $\overline{R}_{\phi,\mathcal{B}}$ is not restricted to $[-C_R,C_R]$, but remains bounded within $[\mu_{\mathrm{reward}}-2C_R,\mu_{\mathrm{reward}}+2C_R]$. We apply this mean constraint only to RewardNet.

\textbf{TaxNet.} TaxNet predicts a sequence-dependent exploitation penalty. For the sequence produced at step $t$, we compute the evolutionary sequence score from Appendix~\ref{app:sequence_probability} and lower-bound its logarithm at $\ell_{\min}=-4.0$:
\begin{equation}
    \ell_{\tau,t}
    =\max\!\left\{
        \log P_{\mathrm{seq}}(\mathbf{s}_{\tau,t+1}),
        \ell_{\min}
      \right\}.
\end{equation}
TaxNet takes $\ell_{\tau,t}$ as input and outputs a scalar tax that can increase for recurrent or SHM-accessible sequence outcomes, thereby discouraging repeated exploitation of a narrow set of evolutionarily likely mutations.
Denoting the learned TaxNet map by $T_\psi$, its scalar output is
\begin{equation}
    \zeta_{\tau,t}=T_\psi(\ell_{\tau,t}).
\end{equation}
Unlike RewardNet, TaxNet is not batch-mean normalized; its output is used directly in the meta-reward.

The meta-reward supplied to the policy is
\begin{equation}
    r_{\mathrm{meta}}^{(\tau,t)}
    =\overline{R}_{\phi,\mathcal{B}}^{(\tau,t)}
    -\zeta_{\tau,t}.
\end{equation}

\textbf{Inner objective.}
Given the learned meta-reward, the mutation policy is optimized with REINFORCE while remaining close to the frozen SHM prior through KL regularization,
\begin{equation}
\begin{aligned}
    \mathcal{L}_{\mathrm{inner}}(\theta;r_{\mathrm{meta}})
    ={}&-\frac{1}{|\mathcal{B}|}
        \sum_{(\tau,t)\in\mathcal{B}}
        G_{\tau,t}
        \log\pi_\theta(u_{\tau,t}\mid\mathbf{s}_{\tau,t})
    \\
    &+\frac{\beta_{\mathrm{KL}}}{|\mathcal{B}|}
        \sum_{(\tau,t)\in\mathcal{B}}
        D_{\mathrm{KL}}\!\left(
            \pi_\theta(\cdot\mid\mathbf{s}_{\tau,t})
            \,\|\,
            \pi_{\mathrm{prior}}(\cdot\mid\mathbf{s}_{\tau,t})
        \right),
\end{aligned}
\end{equation}
For discount factor $\gamma\in[0,1]$, the discounted return is
\begin{equation}
    G_{\tau,t}
    =\sum_{j=t}^{T_\tau-1}
      \gamma^{j-t}r_{\mathrm{meta}}^{(\tau,j)},
\end{equation}
which is used directly in the policy loss. We do not train a value function, subtract a learned state-dependent baseline, or use generalized advantage estimation; consequently, the update has the variance of a Monte Carlo return estimator in addition to the off-policy bias described in Section~\ref{subsec:igc_sel}. Finally, $u_{\tau,t}=(i_{\tau,t},a_{\tau,t})$ denotes the mutation action consisting of the mutation site $i_{\tau,t}$ and the target amino acid $a_{\tau,t}$. Because $r_{\mathrm{meta}}$ depends on $(\phi,\psi)$, the inner objective inherits this dependence through its return.

\textbf{Differentiable inner step.}
RewardNet and TaxNet are optimized indirectly through their effect on policy learning. We therefore perform one differentiable virtual policy update,
\begin{equation}
    \theta'(\phi,\psi)
    =\theta-\eta_{\mathrm{in}}
        \nabla_\theta\mathcal{L}_{\mathrm{inner}}
        \bigl(\theta;r_{\mathrm{meta}}(\phi,\psi)\bigr).
\end{equation}
The computation graph is retained, making $\theta'$ differentiable with respect to $\phi$ and $\psi$ through $r_{\mathrm{meta}}$. The virtual update is used only for meta-optimization and is never applied to the actual policy parameters.

\textbf{Outer meta-objective.}
RewardNet and TaxNet are updated according to the quality of the virtual policy. The objective balances exploration, fidelity to measured affinity, and consistency with the evolutionary sequence-probability score.

Let $H(\pi_{\theta'})$ denote the mean Shannon entropy of the virtually updated position distribution,
\begin{equation}
    H(\pi_{\theta'})
    =-\frac{1}{|\mathcal{B}|}
      \sum_{(\tau,t)\in\mathcal{B}}
      \sum_{i=1}^{L}p_{\theta'}(i\mid\mathbf{s}_{\tau,t})
      \log p_{\theta'}(i\mid\mathbf{s}_{\tau,t}),
    \qquad
    H_{\mathrm{cap}}=\min\{H(\pi_{\theta'}),H_{\max}\}.
\end{equation}

For trajectory $\tau$ of length $T_\tau$, define the log sequence-probability score of its terminal sequence as
\begin{equation}
    q_\tau
    =\log P_{\mathrm{seq}}(\mathbf{s}_{\tau,T_\tau}).
\end{equation}
For each batch element $b=(\tau,t)\in\mathcal{B}$, let
\begin{equation}
\begin{aligned}
    \mathbf{r}_{\mathcal{B}}^{(b)}&=\overline{R}_{\phi,\mathcal{B}}^{(\tau,t)},
    &\mathbf{c}_{\mathcal{B}}^{(b)}&=c_{\tau,t},\\
    \mathbf{q}_{\mathcal{B}}^{(b)}&=q_\tau,
    &\boldsymbol{\ell}_{\mathcal{B}}^{(b)}&=\ell_{\tau,t},\\
    \boldsymbol{\zeta}_{\mathcal{B}}^{(b)}&=\zeta_{\tau,t}.
\end{aligned}
\end{equation}
Thus, the terminal-sequence log score $\mathbf{q}_{\mathcal{B}}^{(b)}$ is shared by all steps from the same trajectory, whereas the resulting-sequence score $\boldsymbol{\ell}_{\mathcal{B}}^{(b)}$ and the remaining quantities are step-specific.

We use the Pearson-correlation fidelity loss
\begin{equation}
\mathcal{L}_{\mathrm{fid}}(x,y) = 1-\rho(x,y), \qquad \rho(x,y) = \frac{\sum_i (x_i-\bar{x})(y_i-\bar{y})} {\sqrt{\sum_i (x_i-\bar{x})^2}\, \sqrt{\sum_i (y_i-\bar{y})^2} +\varepsilon},
\end{equation}
where $\bar{x}$ and $\bar{y}$ are the means of $x$ and $y$, respectively, and $\varepsilon$ is a small constant for numerical stability. The outer objective is
\begin{equation}
\begin{aligned}
\mathcal{L}_{\mathrm{outer}}
\bigl(\phi,\psi;\theta'(\phi,\psi)\bigr)
={}&
-\lambda_H H_{\mathrm{cap}} \\
&+\lambda_R
\Big[
\mathcal{L}_{\mathrm{fid}}(\mathbf{r}_{\mathcal{B}},\mathbf{c}_{\mathcal{B}})
+
\mathcal{L}_{\mathrm{fid}}
\big(
\mathbf{r}_{\mathcal{B}},
\mathbf{q}_{\mathcal{B}}
\big)
\Big] \\
&+\lambda_T\,
\mathcal{L}_{\mathrm{fid}}
\big(
\boldsymbol{\zeta}_{\mathcal{B}},
\boldsymbol{\ell}_{\mathcal{B}}
\big) \\
&+\lambda_{\mathrm{net}}\,
\mathcal{L}_{\mathrm{fid}}
\big(
\mathbf{r}_{\mathcal{B}}-\boldsymbol{\zeta}_{\mathcal{B}},
\mathbf{c}_{\mathcal{B}}
\big).
\end{aligned}
\end{equation}
The entropy term encourages broader exploration. The fidelity terms constrain RewardNet to preserve affinity information and remain consistent with terminal-sequence prevalence, TaxNet to remain aligned with the floored resulting-sequence score, and the combined meta-reward to remain aligned with measured binding.

\textbf{Parameter updates.}
Only the auxiliary-network parameters are updated using the outer objective,
\begin{equation}
    (\phi,\psi)
    \leftarrow
    (\phi,\psi)-\eta_{\mathrm{out}}
    \nabla_{\phi,\psi}
    \mathcal{L}_{\mathrm{outer}}
    \bigl(\phi,\psi;\theta'(\phi,\psi)\bigr).
\end{equation}
This is a total derivative: the entropy term differentiates through the virtual update $\theta'(\phi,\psi)$, while the fidelity terms also provide direct gradients to RewardNet and TaxNet.

The mutation policy is then updated separately using a fresh forward pass with the meta-reward detached from the auxiliary networks,
\begin{equation}
    \theta
    \leftarrow
    \theta-\eta_\theta
    \nabla_\theta
    \mathcal{L}_{\mathrm{inner}}
    \bigl(\theta;\operatorname{stopgrad}(r_{\mathrm{meta}})\bigr).
\end{equation}

Detaching the reward prevents the meta-gradient from providing a second optimization path into the policy. Each training iteration therefore consists of a differentiable virtual update for meta-learning followed by a standard REINFORCE update for the policy.

\section{Hyperparameters}
\label{app:hyperparameters}

Table~\ref{tab:hyperparameters} reports the hyperparameters used for the full RTMG model. The four outer-objective weights are $\lambda_H=3.0$ and $\lambda_R=\lambda_T=\lambda_{\mathrm{net}}=1.0$. RewardNet uses the output cap $C_R=5.0$ and pinned batch mean $\mu_{\mathrm{reward}}=0.9937$, while the entropy contribution is capped at $H_{\max}=4.0$. We set the log sequence-score floor to $\ell_{\min}=-4.0$ and the KL coefficient to $\beta_{\mathrm{KL}}=0.1$.

The differentiable inner update and auxiliary-network outer update both use a learning rate of $10^{-3}$, whereas the deployed mutation policy uses $2\times10^{-4}$. Stage~1 uses batch size 256, and all Stage~2 models use batch size 64. The policy is initialized from the Stage~1 Adapted SHM Prior and uses a six-layer Transformer with hidden dimension 256, eight attention heads, feed-forward dimension 1024, dropout 0.1, and maximum sequence length 149 corresponding to the fixed AHo alignment.

\begin{table}[!hbt]
\centering
\small
\caption{Hyperparameters for the full RTMG model and mutation-policy Transformer.}
\label{tab:hyperparameters}
\begin{tabular}{lll}
\toprule
Component & Hyperparameter & Value \\
\midrule
\multirow{9}{*}{RTMG objective}
& Entropy weight $\lambda_H$ & $3.0$ \\
& Reward fidelity weight $\lambda_R$ & $1.0$ \\
& Tax fidelity weight $\lambda_T$ & $1.0$ \\
& Net-reward fidelity weight $\lambda_{\mathrm{net}}$ & $1.0$ \\
& RewardNet cap $C_R$ & $5.0$ \\
& RewardNet pinned mean $\mu_{\mathrm{reward}}$ & $0.9937$ \\
& Entropy cap $H_{\max}$ & $4.0$ \\
& Log-score floor $\ell_{\min}$ & $-4.0$ \\
& KL coefficient $\beta_{\mathrm{KL}}$ & $0.1$ \\
\midrule
\multirow{5}{*}{Optimization}
& Inner-step learning rate $\eta_{\mathrm{in}}$ & $1\times10^{-3}$ \\
& Outer learning rate $\eta_{\mathrm{out}}$ & $1\times10^{-3}$ \\
& Policy learning rate $\eta_\theta$ & $2\times10^{-4}$ \\
& Stage~1 batch size & $256$ \\
& Stage~2 batch size & $64$ \\
\midrule
\multirow{6}{*}{Policy Transformer}
& Transformer layers & $6$ \\
& Hidden dimension & $256$ \\
& Attention heads & $8$ \\
& Feed-forward dimension & $1024$ \\
& Dropout & $0.1$ \\
& Maximum sequence length & $149$ \\
\bottomrule
\end{tabular}
\end{table}

\section{Compute Cost}
\label{app:compute_cost}

RTMG's bilevel update adds overhead relative to a standard single-level REINFORCE plus KL step because it performs one differentiable inner update to the probe policy $\theta'$, followed by an outer gradient update to RewardNet and TaxNet. We measured this overhead directly using identical hardware, batch size 64, and initialization from the Stage~1 Adapted SHM Prior. One RTMG training step required $0.171\,\mathrm{s}$, compared with $0.121\,\mathrm{s}$ for plain REINFORCE plus KL. Over 140 steps, the corresponding runtimes were $24\,\mathrm{s}$ and $17\,\mathrm{s}$, giving a $1.4\times$ overhead. This is below the $2\times$ overhead suggested by simply counting forward and backward passes.

The lower overhead reflects the asymmetry in parameter counts. The outer gradient updates only RewardNet and TaxNet, which contain $6{,}177$ and $52$ parameters, respectively, for $6{,}229$ parameters in total. This is $769\times$ fewer than the $4{,}788{,}757$ parameters in the mutation policy. Consequently, the second backward pass is inexpensive relative to the additional forward pass and the graph-retaining backward pass implemented with \texttt{create\_graph=True}. RTMG and all Stage~2 baselines use batch size 64, so their reward and diversity comparisons are not confounded by different mini-batch sizes.

\section{Broader Impact and Ethics}

iGC is intended to support antibody research by prioritizing mutation trajectories for subsequent experimental study, but computationally generated candidates should not be interpreted as validated therapeutics or used clinically without laboratory characterization, safety assessment, and regulatory review. Antibody-design methods may also have dual-use implications if applied to harmful targets; responsible use therefore requires target-level biosafety review, controlled experimental validation, and compliance with institutional and legal requirements. This study uses previously collected repertoire, animal, and deep-mutational-scanning datasets and does not recruit human participants or perform new animal experiments.

\section{LLM usage}
During the course of this research, Large Language Models (LLMs) were utilized as assistive tools to enhance both the manuscript and the technical workflow. Specifically, LLMs were employed to polish the academic writing and improve the overall clarity of the text. Furthermore, we leveraged LLMs to assist in configuring our computational experiments and developing the automated data cleaning pipeline, which significantly streamlined our data preprocessing and analysis.

\section{Code and Data Availability}

Code and processed experiment configurations are available at \url{https://github.com/PLACEHOLDER/iGC}. The original Tang and GCReplay datasets are available from their cited sources subject to the corresponding access and usage terms; derived splits and metadata needed to reproduce this work are provided in the repository where redistribution is permitted.

\newpage
\end{document}